\documentclass{article}

\usepackage[preprint]{corl_2023} 

\usepackage{graphicx, float, url, wrapfig}
\usepackage{amsmath, amssymb, bbm, mathtools}
\usepackage{bibunits}
\restylefloat{table}
\usepackage{dsfont} 
\usepackage[subtle]{savetrees}

\usepackage{enumitem}
\usepackage{amsmath, amssymb}
\usepackage{algorithm}
\usepackage{algpseudocode}
\usepackage{booktabs}
\usepackage{multirow}
\usepackage{multicol}

\usepackage[font=footnotesize]{caption}

\newcommand{\bftab}{\fontseries{b}\selectfont}
\usepackage{makecell, tabularx, multirow,  adjustbox}

\usepackage[usenames,dvipsnames]{xcolor}
\usepackage{hyperref}
\hypersetup{
    colorlinks=true,
    linkcolor=MidnightBlue,
    filecolor=OliveGreen,
    urlcolor=MidnightBlue,
    citecolor=MidnightBlue,
    }
\usepackage[capitalise, nameinlink]{cleveref}

\newcommand{\eg}{e.g., }

\newtheorem{proposition}{Proposition}

\newtheorem{claim}{Claim}

\DeclareMathOperator*{\argmax}{arg\!\max}

\newcommand{\olsi}[1]{\,\overline{\!{#1}}} 

\usepackage{xspace}
\newcommand{\knowno}{\textsc{KnowNo}\xspace}

\newif\ifarxiv
\arxivtrue

\title{
Robots That Ask For Help: Uncertainty Alignment \\ for Large Language Model Planners
}

\author{
  Allen Z. Ren$^{1,2}$, Anushri Dixit$^1$, Alexandra Bodrova$^1$, Sumeet Singh$^2$, Stephen Tu$^2$, \\ \bf{Noah Brown}$^2$, Peng Xu$^2$, Leila Takayama$^2$, Fei Xia$^2$, Jake Varley$^2$, Zhenjia Xu$^2$,  \\ \bf{Dorsa Sadigh}$^2$, Andy Zeng$^2$, Anirudha Majumdar$^{1,2}$
\\
  $^1$Princeton University, $^2$Google DeepMind\\
}

\begin{document}
\maketitle

\setlength{\abovedisplayskip}{3pt}
\setlength{\belowdisplayskip}{3pt}
\setlength{\abovedisplayshortskip}{3pt}
\setlength{\belowdisplayshortskip}{3pt}


\vspace{-7pt}
\begin{abstract}
Large language models (LLMs) exhibit a wide range of promising capabilities --- from step-by-step planning to commonsense reasoning --- that may provide utility for robots, but remain prone to \emph{confidently} hallucinated predictions. In this work, we present \knowno, which is a framework for measuring and aligning the uncertainty of LLM-based planners such that they \emph{know when they don't know} and ask for help when needed. \knowno builds on the theory of \emph{conformal prediction} to provide statistical guarantees on task completion while minimizing human help in complex multi-step planning settings. 
Experiments across a variety of simulated and real robot setups that involve tasks with different modes of ambiguity (\eg from spatial to numeric uncertainties, from human preferences to Winograd schemas) show that \knowno performs favorably over modern baselines (which may involve ensembles or extensive prompt tuning) in terms of improving efficiency and autonomy, while providing formal assurances. 
\knowno can be used with LLMs out of the box without model-finetuning, and suggests a promising lightweight approach to modeling uncertainty that can complement and scale with the growing capabilities of foundation models.\footnote{Webpage with additional information, videos, and code: 
\url{https://robot-help.github.io}}
\end{abstract}
\vspace{-10pt}

\section{Introduction}
\label{sec:intro}



How can we endow our robots with the ability to \emph{know when they don't know}? Accurately modeling and accounting for uncertainty is a longstanding challenge towards robots that operate reliably in unstructured and novel environments. In this work, we study this challenge in the context of \emph{language-instructed} robots. Language provides a natural and flexible interface for humans to specify tasks, contextual information, and intentions, while also allowing us to provide help and clarification to robots when they are uncertain. 

Recently, approaches that leverage large language models (LLMs) for planning \citep{ahn2022can, huang2022inner} have demonstrated the ability to respond to natural and unstructured language instructions to generate temporally extended plans. These approaches enable leveraging the vast amount of prior knowledge and rich context embedded in pretrained LLMs, and lead to substantial abstract reasoning capabilities. However, one of the major challenges with current LLMs is their tendency to \emph{hallucinate}, i.e., to \emph{confidently} generate outputs that are plausible but incorrect and untethered from reality. Such false confidence in incorrect outputs poses a significant challenge to LLM-based planning in robotics. Moreover, natural language instructions in real-world environments often contain a high degree of \emph{ambiguity} inherently or unintentionally from humans, and confidently following an incorrectly constructed plan could lead to undesirable or even unsafe actions. 

\begin{figure}
\centering\includegraphics[width=1\linewidth]{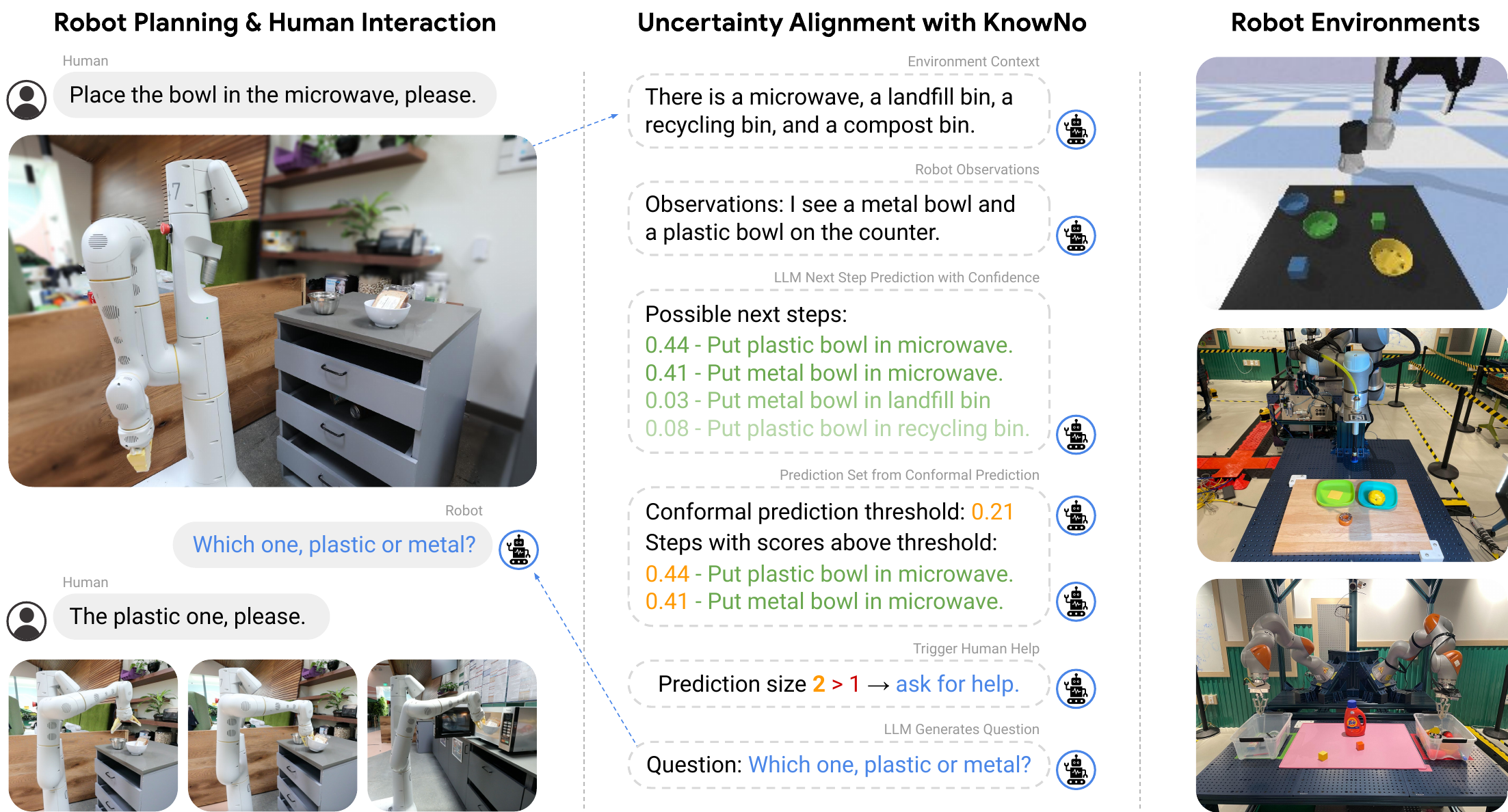}
    \vspace{-2pt}
    \caption{\knowno uses Conformal Prediction (CP) to align the uncertainty of LLM planners. Given a language instruction, an LLM generates possible next steps and its confidences (scores) in these options. CP then provides a prediction set that includes the options with scores above a certain quantile. If there is more than one option in the set, the robot asks for help. Experiments across multiple embodiments and a variety of ambiguous situations show that \knowno significantly improves efficiency and autonomy compared to baselines.}
    \label{fig:teaser}
    \ifarxiv \vspace{-15pt} \else \vspace{-20pt} \fi
\end{figure}

As an example (\cref{fig:teaser}), a robot tasked with heating food may be asked to ``place the bowl in the microwave''; if there are multiple bowls on the counter, the instruction is ambiguous. Moreover, the metal bowl may not be safe for the microwave. Rather than acting in this ambiguous setting and damaging the microwave or even causing a fire, the robot should  know when it doesn't know and ask for clarification instead (e.g., ask which bowl should be placed in the microwave).
Prior work in language-based planning either does not seek such clarifications \citep{ahn2022can} or does so via extensive prompting \citep{huang2022inner}, which requires careful prompt engineering to prevent the robot from excessively relying on seeking assistance. Moreover, prior approaches do not provide a way to ensure that asking for help results in a desired level of task success. We formalize these challenges via two desiderata: (i) \emph{calibrated confidence:} the robot should seek sufficient help to ensure a statistically guaranteed level of task success specified by the user, and (ii) \emph{minimal help:} the robot should minimize the overall amount of help it seeks by narrowing down possible ambiguities in a task.
We collectively refer to these sufficiency and minimality conditions as \emph{uncertainty alignment}.

{\bf Statement of contributions.} We propose \knowno --- \emph{Know When You Don't Know} --- a framework for aligning the uncertainty of LLM-based planners utilizing the theory of \emph{conformal prediction (CP)}~\citep{vovk2005algorithmic, angelopoulos2023conformal}. We make the following contributions:
    \textbf{(1)} Given a language instruction, we utilize a pre-trained LLM with uncalibrated confidence to generate a \emph{set} of possible actions for the robot to execute next. We demonstrate how to use CP to select a subset of these options, which allows the robot to decide an action to execute (if the subset is a singleton) or to ask for help otherwise. 
    \textbf{(2)} We prove theoretical guarantees on calibrated confidence in both single-step and multi-step planning problems: with a user-specified level $1-\epsilon$, the robot performs the tasks correctly in $1-\epsilon \ \%$ of scenarios by asking for help when it deems it necessary. CP also minimizes the average size of prediction sets, thus addressing the goal of minimal help.
    \textbf{(3)} We evaluate \knowno in both simulation and hardware with a suite of language-instructed manipulation tasks with various types of potential ambiguities (e.g., based on spatial locations, numerical values, attributes of objects, and Winograd schemas). Experiments across multiple settings and embodiments validate the ability of \knowno to provide statistically guaranteed levels of task success while 
    reducing the amount of help required 
    by $10-24\%$ as compared to baseline approaches.

\section{Overview: Robots that Ask for Help}
\label{sec:overview}

\paragraph{Language-based planners.} Language model planners can generate step-by-step robot plans, where each step $y$ is composed of variable-length sequences of symbols $(\sigma_1,\sigma_2,\dots,\sigma_k)$, \eg text tokens as input to a language-conditioned policy \cite{ahn2022can} (see \cref{fig:teaser}), or robot code executed by an interpreter \cite{liang2022code}. Pretrained autoregressive LLMs predict each step $y$, whose joint probability over tokens can be factorized as the product of conditional probabilities of next token prediction $p(y)=\prod_{i=1}^{k} p(\sigma_i \mid \sigma_1, \ldots, \sigma_{i-1})$. Here, we are interested in characterizing the uncertainty of next step prediction $p(y)$. The distribution of $p$ remains highly sensitive to variable-length $k$; hence $p(y)$ on its own serves as a rather poor scoring function \cite{murray2018correcting} particularly when steps in a plan are expressed in natural language (our experiments in \cref{app:discussions} also show that using $p(y)$ directly for calibration leads to poor performance).

\begin{figure}[t]
    \vspace{-5pt}
    \centering\includegraphics[width=1.0\linewidth]{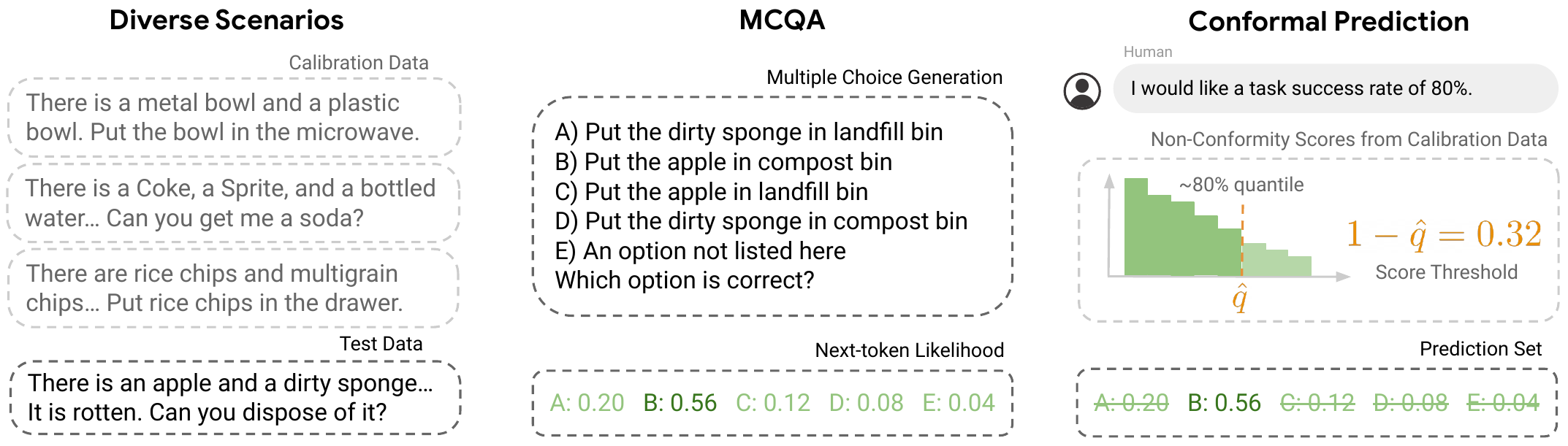}
    \caption{\knowno formulates LLM planning as MCQA by first prompting an LLM to generate plausible options, and then asking it to predict the correct one. Based on the next-token likelihoods from a calibration dataset, CP finds the quantile value $\hat{q}$ such that all options with a score $\geq 1-\hat{q}$ are included in the prediction set in a test scenario. The resulting sets are guaranteed to cover the true option with the user-specified probability.}
    \label{fig:mcqa-cp}
    \ifarxiv \vspace{-15pt} \else \vspace{-20pt} \fi
\end{figure}

\paragraph{Planning as multiple-choice Q\&A.} We can address this length bias with a simple trick. First, with a few-shot prompt that includes possible next steps in a few scenarios (\cref{fig:app-mc-pre-prompt}), the LLM generates a \emph{set} $\{y^i\}$ of candidate next steps (e.g., ``Put plastic bowl in microwave", ``Put metal bowl in microwave", etc., in \cref{fig:teaser}) that are \emph{semantically different}. Then the task of choosing among them is formatted as multiple-choice Q\&A (MCQA). This eliminates plans that the LLM considers unlikely and reduces the problem of next-step prediction down to a single next-token prediction --- aligning with LLM log-likelihood loss functions and LLM training data (\eg MCQA datasets \cite{srivastava2022beyond, hendrycks2020measuring}). These probabilities can serve as normalized scores that can be used by various uncertainty quantification methods such as thresholding and ensemble methods. In this work, we use these normalized scores within a conformal prediction (CP) framework.  Specifically, CP uses a held-out calibration set of example plans in different scenarios to generate a reduced prediction set of plans among $\{y^i$\} (\cref{fig:mcqa-cp}). The LLM is certain if this prediction set is a singleton, and triggers help from a human otherwise. 
\cref{app:mcqa} details additional rationale of applying MCQA to evaluate the \emph{semantic uncertainty} of the LLM.
\paragraph{Robots that ask for help.} In this work, we show that LLM planning --- combined with CP for uncertainty estimation --- can effectively enable robots to interact with an environment, and ask for help when needed. The environment $e$ can be formulated as a partially observable Markov decision process (POMDP): at any given state $s^t$ at time $t$, given a user instruction $\ell$, the robot executes an action $a^t$ according to a policy $\pi$, then transitions to a new state $s^{t+1}$. Our policy $\pi$ is composed of four parts (\cref{fig:teaser}):
\begin{enumerate}
[leftmargin=*,topsep=-1pt,itemsep=-2pt,partopsep=1ex,parsep=1ex]
\item Multiple-choice generation: An LLM generates a diverse set of candidate plans labeled with `$A$', `$B$', `$C$', `$D$'
, and an additional possible plan, `$E) \ \text{an option not listed here}$', which is appended post-hoc. We denote the set of labels by $\mathcal{Y} \coloneqq \{\textrm`A\textrm', \textrm`B\textrm', \textrm`C\textrm', \textrm`D\textrm', \textrm`E\textrm'\}$. These plans are generated by prompting the LLM with \emph{context} $x^t$, which is text that includes (1) the robot observation at each time step (e.g., using a vision-based object detector or an oracle; see \cref{fig:teaser}), (2) the user instruction, and (3) few-shot examples of possible plans in other scenarios. An \emph{augmented context} $\tilde{x}^t$ is obtained by appending the LLM-generated plans to the context $x^t$.
\item Prediction set generation: We use CP to choose a \emph{subset} $C(\tilde{x}^t) \subseteq \mathcal{Y}$ of candidate plans using the LLM's (uncalibrated) confidence $\hat{f}(\tilde{x}^t)_y$ in each prediction $y \in \mathcal{Y}$ given the context $\tilde{x}^t$.
\item Human help: If the prediction set is a non-singleton, the robot leverages help from a human (or any other supervisor agent, denoted as a function $f_\mathcal{H}$)
to arrive at an unambiguous next step $y_\mathcal{H}\in C(\tilde{x}^t)$.
\item Low-level control: A low-level module $\varphi$ converts the plan in $y_\mathcal{H}$ to an action $a^t = \varphi(y_\mathcal{H})$.
\end{enumerate}


\paragraph{Goal: uncertainty alignment.} Often in real-world settings, language instructions $\ell$ can be ambiguous, \eg ``place the bowl in the microwave'' does not specify that the human means the plastic bowl (\cref{fig:teaser}). Our goal in this work is to address uncertainty alignment: achieve a desired level of task success while minimizing human help. We formalize this by considering a joint distribution $\mathcal{D}$ over \emph{scenarios} $\xi \coloneqq (e,\ell,g)$, where $e$ is an environment (POMDP), $\ell$ is a (potentially ambiguous) language instruction, and $g$ is a goal (e.g., formulated as a subset of acceptable states in the POMDP and partially observable through $l$). Importantly, we do not assume knowledge of $\mathcal{D}$, except that we can sample a finite-size dataset of i.i.d. scenarios from it. We formalize uncertainty alignment in our setting as (i) \emph{calibrated confidence:} the robot's policy (with human help as described above) succeeds with a user-specified probability $1-\epsilon$ over new scenarios $\xi \sim \mathcal{D}$, and (ii) \emph{minimal help:} the policy minimizes the number $|C(\cdot)|$ of options presented to the human on average across scenarios $\xi \sim \mathcal{D}$.

\section{Calibrating LLM Confidence with Conformal Prediction}
\label{sec:approach}

The MCQA setup above allows us to apply CP to obtain calibrated confidence guarantees while (approximately) minimizing help. We introduce CP below, and then present the different practical settings we consider (possibly involving multiple planning steps and/or multiple correct plans per step).

\subsection{Background: Conformal Prediction}
\label{subsec:cp}

For now, we drop the timestep superscript and consider a generic MCQA setup with pairs $(\tilde{x}, y)$ consisting of input $\tilde{x}$  and true label $y$. Suppose there is a \emph{calibration} set $Z = \{z_i = (\tilde{x}_i, y_i)\}_{i=1}^N$ of such pairs drawn i.i.d. from an \emph{unknown} distribution $\mathcal{D}$ over $\mathcal{Z}:= \mathcal{X} \times \mathcal{Y}$. Now, given a new i.i.d. sample $z_\text{test} = (\tilde{x}_\text{test}, y_\text{test})$ with unknown true label $y_\text{test}$, CP generates a \emph{prediction set} $C(\tilde{x}_\text{test}) \subseteq \mathcal{Y}$ that contains $y_\text{test}$ with high probability \citep{vovk2005algorithmic}:
\begin{equation}
\label{eq:marginal_coverage}
    \mathbb{P}\big(y_\text{test} \in C(\tilde{x}_\text{test})\big) \geq 1-\epsilon,
\end{equation}
where $1-\epsilon$ is a user-specified value (desired task success level in our setting) that affects the size of $C(\cdot)$.

To generate $C(\tilde{x}_\text{test})$, 
CP first uses the LLM's confidence $\hat{f}$ (cf. \cref{sec:overview}) to evaluate the set of \emph{nonconformity scores} $\{\kappa_i = 1 - \hat{f}(\tilde{x}_i)_{y_i}\}_{i=1}^N$ over the calibration set --- the higher the score is, the less each data in the calibration set \emph{conforms} to the data used for training $\hat{f}$. 
Then CP performs calibration by defining $\hat{q}$ to be the $\frac{\lceil (N+1)(1-\epsilon) \rceil}{N}$ empirical quantile of $\kappa_1, \dots, \kappa_N$. Lastly, CP generates $C(\tilde{x}_\text{test}) = \{y \in \mathcal{Y} \ | \ \hat{f}(\tilde{x}_\text{test})_y \geq 1-\hat{q})\}$, i.e., the prediction set that includes all labels that the predictor is at least $1-\hat{q}$ confident in. The generated prediction set ensures that the coverage guarantee in \cref{eq:marginal_coverage} holds. 


\textbf{Dataset-conditional guarantee.} The probability in \cref{eq:marginal_coverage} is over \emph{both} the sampling of the calibration set $Z$ and $z_\text{test}$ (i.e., a \emph{marginal} guarantee).
Thus, to ensure the desired probability of coverage for each new $z_\text{test}$, one needs a fresh calibration set.
Instead, we apply the following \emph{dataset-conditional} guarantee \citep{vovk2012conditional} which is conditioned on a particular calibration dataset being sampled, and thus can be applied to new test data without re-calibration:
\begin{equation}
\label{eq:conditional_coverage}  \mathbb{P}\big(y_\text{test} \in C(\tilde{x}_\text{test}) \ | \ \{z_1, \dots, z_N\}\big) \geq \text{Beta}_{N+1-v, v}^{-1}(\delta), \quad v := \lfloor (N+1)\hat{\epsilon} \rfloor,
\end{equation}
where $\text{Beta}_{N+1-v, v}^{-1}(\delta)$ denotes the inverse CDF (quantile) level of $\delta$ in a Beta distribution with parameters $N+1-v$ and $v$, and $\hat{\epsilon}$ is the threshold used for calibration. In practice, we use a modest-sized calibration dataset ($N=400$) and $\delta=0.01$, and adjust $\hat{\epsilon}$ to achieve the desired $1-\epsilon$ coverage (with probability $1-\delta = 0.99$ over the sampling of the calibration set). 

\textbf{Minimal prediction set size.} From \citep[Thm.~1]{sadinle2019least}, $C(\cdot)$ achieves the smallest average set size among possible prediction schemes $\mathcal{C}$ that achieve the coverage guarantee, if $\hat{f}(\tilde{x})_y$ models true conditional probabilities:
\begin{equation}
\label{eq:minimality}
    \min_{C \in \mathcal{C}} \ \underset{(\tilde{x},\cdot) \sim \mathcal{D}}{\mathbb{E}} \big[|C(\tilde{x})|\big], \ \text{subject to} \ \eqref{eq:marginal_coverage}.
\end{equation} 
The assumption that $\hat{f}$ models true conditional probabilities may be a good approximation for LLMs trained on large-scale data with a proper scoring rule \cite{jiang2021can}; one can also obtain bounds on near-optimal average set size for CP using $\hat{f}$ that approximately models conditional probabilities \cite{bai2022efficient, sadinle2019least}, but we omit these results for brevity. We emphasize that the CP coverage guarantees hold regardless of the accuracy of $\hat{f}$. Overall, CP is a powerful and easy-to-use statistical tool to produce (1) tight coverage guarantees---addressing the goal of \emph{calibrated confidence}, and (2) small prediction sets for unseen data given a blackbox predictor like an LLM and an unknown data distribution---addressing our second goal of \emph{minimal help}.


\subsection{Single-Step Uncertainty Alignment}
\label{subsec:single-step}

We now demonstrate how to use CP to achieve uncertainty alignment for LLM-based planning 
with a user-specified task completion rate $1-\epsilon$. We first consider a single-step setting, where the LLM plans only once given a context. For simplicity, we again drop the timestep superscript $t$ in this section. 


\paragraph{Data collection.}  We collect $N$ i.i.d. scenarios from the distribution $\mathcal{D}$, and the corresponding contexts summarizing the robot observation and instruction (\cref{sec:overview}). We use the MCQA approach from~\cref{sec:overview} to generate candidate plans and then label each augmented context $\tilde{x}$ (i.e., context combined with plans) with the correct label (here and in \cref{sec:multi-step}, we assume that there is a \emph{unique} correct candidate plan; we provide an extension to multiple acceptable options in \cref{app:multi-option}). We thus obtain a calibration set $Z = \{z_i = (\tilde{x}_i, y_i)\}_{i=1}^N$ with pairs of augmented contexts and correct labels.

\paragraph{Calibration.} Next we follow \cref{subsec:cp} to perform calibration: first adjust $\hat{\epsilon}$ to achieve the $1-\epsilon$ coverage based on~\cref{eq:conditional_coverage} and then find the quantile $\hat{q}$. Given a new context $\tilde{x}_\text{test}$ (after MCQA in a new scenario) at test time, we can construct the calibration set $C(\tilde{x}_\text{test})$ that contains $y_\text{test}$ with $1-\epsilon$ probability.

\paragraph{Triggering help.} If $C(\tilde{x}_\text{test})$ is a singleton, the robot executes the corresponding plan. Otherwise, we deem the LLM uncertain over possible actions and trigger human help. The robot presents the human with $C(\tilde{x}_\text{test})$ (including the corresponding plans in text) and asks the human to choose one\footnote{In practice we convert the prediction set to a question in natural language (\cref{app:implementation}).}. The human chooses $y_\text{test}$ if $y_\text{test} \in C(\tilde{x}_\text{test})$\footnote{If the correct option in $C(\tilde{x}_\text{test})$ is `E', the human provides the correct action that was not listed by the robot. 
}, or halts the operation otherwise. This setup turns the coverage guarantee from CP to the task completion guarantee:

\vspace{-3pt}
\begin{proposition}[Single-step uncertainty alignment]\label{proposition:single_step}
Consider a single-step setting where we use CP with coverage level $1-\epsilon$ to generate prediction sets and seek help whenever the set is not a singleton. With probability $1-\delta$ over the sampling of the calibration set, the task completion rate over new test scenarios drawn from $\mathcal{D}$ is at least $1-\epsilon$. If $\hat{f}$ models true conditional probabilities, the average prediction set size is minimized among possible prediction schemes that achieve $1-\epsilon$ completion rate.
\end{proposition}
\vspace{-3pt}

The proof immediately follows from the fact that under the assumption of accurate human help, the robot fails only when the prediction set does not contain the true label;
the prediction set minimality follows from \cref{eq:minimality}.
Thus, our approach addresses the goals of calibrated confidence and minimal help from \cref{sec:overview}.

\subsection{Multi-Step Uncertainty Alignment}
\label{sec:multi-step}

Now we extend the CP-based uncertainty alignment approach to settings where the LLM plans in multiple timesteps. This setting can be helpful when the LLM receives feedback from the environment or human between steps. However, the original CP formulation cannot be applied here since the context $x^t$ between steps are dependent; moreover, the robot's actions at step $t$ influence the distribution over contexts that the robot observes at future steps. Thus, the i.i.d. assumption for the coverage guarantee is no longer valid. Here, we present a novel extension of CP to multi-step settings that tackles this challenge.

\paragraph{Sequence-level calibration.} The key ideas are to (i) \emph{lift} the data to sequences, and (ii) perform calibration at the sequence level using a carefully designed nonconformity score function that allows for \emph{causal} reconstruction of the prediction set at test time. Suppose that each data point consists of a sequence of augmented context $\olsi{x} = (\tilde{x}^0, \tilde{x}^1, \dots, \tilde{x}^{T-1})$ and true labels $\olsi{y} = (y^0, y^1, \dots, y^{T-1})$, where $T$ is the time horizon and $\tilde{x}^t$ arises from having performed the \emph{correct} actions in previous steps. The distribution $\mathcal{D}$ over scenarios induces a distribution over data sequences. We can again collect a calibration set $\olsi{Z} = \{\olsi{z}_i = (\olsi{x}_i, \olsi{y}_i)\}_{i=1}^N$. Next we use the lowest score over the timesteps as the score for the sequence\footnote{We overload notation here and use $\hat{f}$ to also assign confidence scores to sequences.}:
\begin{equation}
    \hat{f}(\olsi{x})_{\olsi{y}} := \min_{t \in [T]} \ \hat{f}(x^t)_{y^t}.
\end{equation}
With the standard calibration procedure in \cref{subsec:cp}, we construct a \emph{sequence-level} prediction set $\olsi{C}(\olsi{x}_\text{test}) := \{ \olsi{y} \in \mathcal{Y}^T \ | \ \hat{f}(\olsi{x}_\text{test})_{\olsi{y}} \geq 1-\hat{q} \}$
for a new context sequence $\olsi{x}_\text{test}$ with the quantile $\hat{q}$.

\paragraph{Causal reconstruction of $C(\olsi{x})$ at test time.} Note that $\olsi{C}(\olsi{x}_\text{test})$ is constructed with the full sequence $\olsi{x}_\text{test}$ at once.
However, at test time, we do not see the entire sequence of contexts all at once but rather $x_\text{test}^t$ one at a time. We thus need to reconstruct
$\olsi{C}(\olsi{x}_\text{test})$ in a \emph{causal} manner (i.e., always relying only on current and past information). Consider the causally constructed prediction set $
C^t(x_\text{test}^t) := \{ y^t \ | \ \hat{f}(x_\text{test}^t)_{y^t} \geq 1-\hat{q} \}$ at time $t$ using the same quantile level $\hat{q}$ from the non-causal calibration above, and define $C(\olsi{x}_\text{test}) := C^0(x_\text{test}^0) \times C^1(x_\text{test}^1) \times \dots \times C^{T-1}(x_\text{test}^{T-1})$. We would like to obtain a lower bound on the sequence-level coverage: $\mathbb{P}(\olsi{y}_\text{test} \in C(\olsi{x}_\text{test})) \geq 1-\epsilon$.


\vspace{-3pt}
\begin{claim}\label{claim:multi_CP} For any $\olsi{y} \in \mathcal{Y}^T$, $\olsi{y} \in \olsi{C}(\olsi{x}_\text{test}) \iff \olsi{y}
\in C(\olsi{x}_\text{test})$.
\end{claim}
\vspace{-3pt}

\begin{proposition}[Multi-step uncertainty alignment]\label{proposition: multi_CP}
Consider a multi-step setting where we use CP with coverage level $1-\epsilon$ to causally construct the prediction set and seek help whenever the set is not a singleton at each timestep. With probability $1-\delta$ over the sampling of the calibration set, the task completion rate over new test scenarios drawn from $\mathcal{D}$ is at least $1-\epsilon$. If $\hat{f}$ models true conditional probabilities, the average prediction set size is minimized among possible prediction schemes that achieve $1-\epsilon$ completion rate.
\end{proposition}
The proofs are deferred to \cref{app:multi-step}. Claim 1 allows us to construct causal prediction sets from non-causal calibration. We then show that the sequence-level task completion rate guarantee still holds. 

\paragraph{Multiple acceptable options.} Often, there can be multiple acceptable options at the same timestep, e.g., the task is to bring the human a soda, and either the Coke or Sprite on the table is acceptable. In such settings, we would like the prediction set to contain \emph{at least one} acceptable option. We extend our method and confidence guarantees to this setting for both single- and multi-step problems in \cref{app:multi-option} and \cref{app:multi-step-multi-option}. 

\section{Experiments}
\label{sec:experiments}

We evaluate our framework in a diverse set of language-instructed tasks and environments below, and demonstrate its effectiveness in achieving a user-specified task completion rate while minimizing user help. 
We use PaLM-2L \citep{palm2} as the LLM in all examples unless otherwise noted. 

\paragraph{Scenario Distribution and Calibration Dataset.} \knowno can be applied to \emph{unknown} scenario distribution if one can collect i.i.d. samples from it for calibration. In future deployment, we envision that a robot can interact with an end-user (e.g., in a home) to collect a dataset through interactions with the human; the human can provide their true intentions during this data collection phase. For experiments here, we assume that the human faithfully provides help. We parameterize the scenario distribution for each experiment with details shown in \cref{app:experiment}. Each calibration dataset is then generated by random sampling from the corresponding distribution. Labeling the calibration data takes about 4 hours (for 400 examples) in the multi-step setting and 1.5 hours in single-step settings.

\paragraph{Baselines.}
A straightforward way to construct prediction sets given a desired $1-\epsilon$ coverage is to rank options according to confidence and construct a set such that the cumulative confidence exceeds $1-\epsilon$; we consider two baselines that are based on such cumulative thresholding but use different kinds of scores: \textbf{Simple Set} uses the same $\hat{f}$ as \knowno; \textbf{Ensemble Set} \citep{lu2021ensemble} instead uses the frequencies of the LLM outputting $y \in \mathcal{Y}$ (out of 20 trials total) with randomized sampling of few-shot examples in the prompt.
However, the resulting prediction sets
are not guaranteed to achieve $1-\epsilon$ coverage as the probabilities can be miscalibrated \citep{guo2017calibration}, and often include additional unnecessary options \citep{angelopoulos2020uncertainty}.
Instead of using cumulative thresholding, \knowno constructs prediction sets by including options with scores higher than a threshold computed using CP, which results in statistical guarantees.
We also introduce two prompt-based baselines: \textbf{Prompt Set} prompts the LLM to directly output the prediction set (\eg ``Prediction set: [A, C]''); \textbf{Binary} prompts the LLM to directly output a binary indicator for uncertainty (\eg ``Certain/Uncertain: Certain''), which is used in other LLM-based planning work \citep{huang2022inner} for triggering human intervention. Note that the $\epsilon$ level is not used in Prompt Set or Binary, and so the user cannot explicitly control the task success rate. Lastly, we consider \textbf{No Help} where the option with the highest score is always executed without any human intervention.






\subsection{Simulation: Tabletop Rearrangement}
\label{subsec:sim}

\begin{wrapfigure}[12]{r}{0.4\textwidth}
\vspace{-37pt}
\centering
\includegraphics[width=0.4\textwidth]{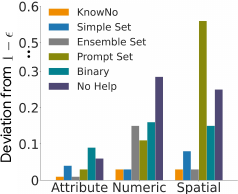}
\caption{Deviation from specified task success level $1-\epsilon = 0.85$ to the empirical success rate for the three settings in \textbf{Simulation}. 200 trials are run for each method/setting.}
\label{fig:coverage}
\end{wrapfigure}


A robot arm is asked to rearrange objects on a table in the PyBullet simulator \citep{coumans2020} (\cref{fig:teaser} right top).
Each scenario is initialized with three bowls and blocks of green, yellow, and blue colors. The task is to move a certain number of blocks or bowls towards a different object or at a specific location around it. We introduce three settings based on different types of ambiguities in the user instruction: (1) \emph{Attribute} (e.g., referring to the bowl with the word ``receptacle''), (2) \emph{Numeric} (e.g., under-specifying the number of blocks to be moved by saying ``a few blocks''), and (3) \emph{Spatial} (e.g., ``put the yellow block next to the green bowl'', but the human has a preference over placing it at the front/back/left/right). For each setting, we construct a distribution over scenarios (detailed in \cref{app:experiment}) and perform experiments separately. This is a single-step setting with single acceptable option per step.

\paragraph{\knowno achieves target task success rate consistently.} First, we investigate whether \knowno and the baselines achieve a given target task success rate \emph{consistently} in the three settings --- we set the failure level $\epsilon = 0.15$. In \cref{fig:coverage} we show the difference between achieved and target rates for all methods. Results show that \knowno achieve the least deviations overall, due to the coverage guarantee from CP. Simple Set and Ensemble Set cannot achieve coverage consistently. Prompt Set, Binary, and No Help have larger deviations from the target since the user has no control over the error rate. Also, as the scenarios get increasingly ambiguous (least in Attribute and most in Spatial), the baselines show larger deviations.

\begin{wrapfigure}[15]{r}{0.65\textwidth}
\vspace{-20pt}
\begin{center}
\includegraphics[width=0.65\textwidth]{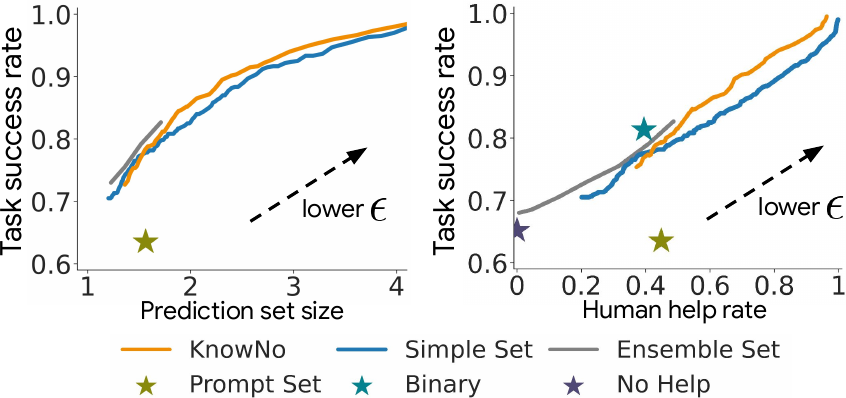}
\caption{Comparison of task success rate vs average prediction set size (Left) and vs. human help rate (Right) in \textbf{Simulation} averaged over the three settings. 200 trials are run for each method. $\epsilon$ is varied from 0.25 to 0.01 for \knowno, and from 0.6 to 0.01 for Simple Set and Ensemble Set. Binary and No Help are not shown on the left since prediction sets are not provided.}
\label{fig:sim-comparisons}
\vspace{-10pt}
\end{center}
\end{wrapfigure}

\paragraph{\knowno achieves high task success rate with lower human help as $\epsilon$ varies.} In \cref{fig:sim-comparisons} we vary the target error rate $\epsilon$ and show the curves of task success rate vs. prediction set size and human help rate averaged over the three settings. For \knowno, Simple Set, and Ensemble Set, specifying a lower $\epsilon$ improves the empirical task success rate while also requiring more human help. The most natural comparison is between \knowno and Simple Set, as both use next-token probabilities from the LLM as the confidence score.
\knowno achieves higher success-to-help ratios across $\epsilon$ levels, thanks to calibrated confidence from CP.
Meanwhile, Prompt Set and Binary do not allow controlling success rates. Prompt Set performs the worst, indicating the challenge of prompting-based methods for calibrated prediction sets. Binary works favorably at some success levels, but lacks flexibility and doesn't provide prediction sets for human feedback. In addition, \cref{fig:app-sim-comparisons-setting} shows the results for individual ambiguity settings. As the scenarios become more ambiguous, \knowno shows a greater reduction of human help compared to Simple Set --- as much as 24\% at certain success levels.

\paragraph{Ensemble Set can perform well but is computationally expensive.} \cref{fig:sim-comparisons} also shows that Ensemble Set provides high task success rate with small amount of human help at higher $\epsilon$ levels. However, there are two main drawbacks. First, we find in some scenarios that even with 20 randomized prompts, the LLM can fail to choose the correct option and thus assigns zero probability to it. As shown in \cref{fig:sim-comparisons}, this means that Ensemble Set can fail to improve once it reaches some level of human help. Second, it requires 20$\times$ inference time compared to other methods. Investigating how to lower the computational burden and combining ensemble-based probabilities with CP can be a fruitful future direction.

\subsection{Hardware: Multi-Step Tabletop Rearrangement}

\begin{figure}[h]
\begin{center}
\includegraphics[width=1.0\textwidth]{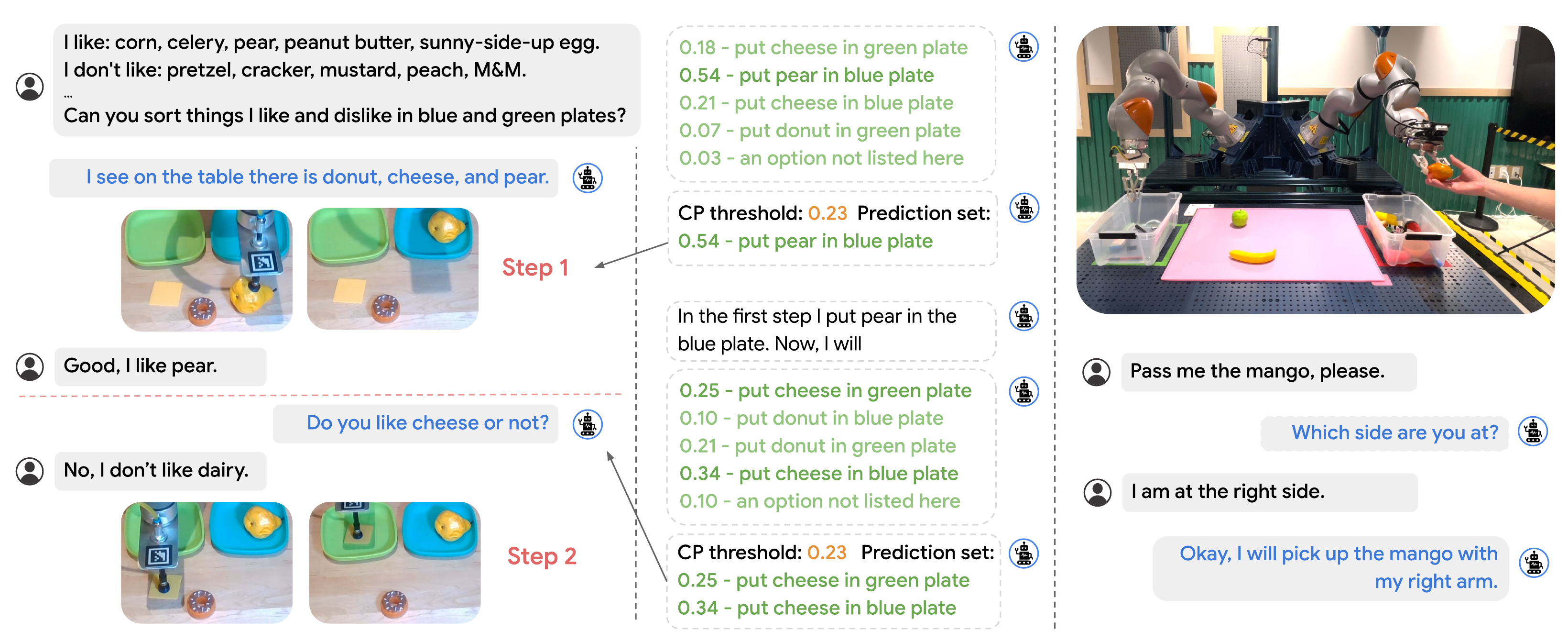}
\caption{(Left) Multi-step CP is applied in \textbf{Hardware Tabletop Rearrangement}. (Right) CP models ambiguity in possible human locations and triggers clarification from the human in \textbf{Bimanual}.}
\vspace{-20pt}
\label{fig:multi-step-prompt}
\end{center}
\end{figure}

In this example, a UR5 robot arm is asked to sort a variety of toy food items on a table (\cref{fig:multi-step-prompt} left). In each scenario, three items are placed on the table initially, and the task is to sort them based on human preferences; we simulate a human with strong preferences for healthy food like eggs and fruits, and dislike for less healthy food like donuts and Skittles candies. To introduce ambiguities, the context for the LLM reveals only a subset of the preferences. Here we consider a multi-step setting with possibly multiple acceptable options per step --- the LLM plans the new step conditioned on the previous action taken. 

\setlength{\tabcolsep}{1pt}
\begin{wraptable}[6]{R}{6.5cm}
\vspace{-35pt}
\centering
\scriptsize
\begin{tabular}{ccccccc}\\\toprule
Method & $1-\epsilon$ & Plan Succ & Task Succ & Set Size & Help-Step & Help-Trial \\\midrule
\knowno & 
    0.75 & 0.76 & 0.74 & \bftab 1.72 & \bftab 0.58 & \bftab 0.92 \\
Simple Set &
    0.58 & 0.76 & 0.72 & 2.04 & 0.72 & 1.00 \\ 
No Help &
    - & 0.41 & 0.38 & - & 0 & 0 \\ 
\bottomrule
\end{tabular}
\caption{Results for \textbf{Hardware Multi-Step Tabletop Rearrangement}. Plan success rate is fixed between \knowno and Simple Set  for comparing the other metrics.
}
\label{tab:tabletop-multi-step}
\end{wraptable}


\paragraph{\knowno reduces step-wise and trial-wise intervention rates in multi-step setting.} 
Since \cref{subsec:sim} has shown that Ensemble Set can be expensive (even more so in the multi-step setting) and Prompt Set and Binary can fail to achieve the user-specified success level, we focus on comparing \knowno with Simple Set for the remainder of the evaluation. Here we set the desired error level $\epsilon=0.25$. Since Simple Set does not provide calibrated coverage, we first find $\epsilon=0.42$ for Simple Set to achieve the same planning error rate as \knowno in simulation. Then we run 50 trials for both methods in hardware. \cref{tab:tabletop-multi-step} shows that \knowno reduces the human help rate by 14\% step-wise and 8\% trial-wise, while also reducing the average prediction set size. Compared to Simple Set which uses a much higher $\epsilon$, \knowno achieves the specified \emph{trial-level} task success rate precisely by leveraging the Multi-Step Uncertainty Alignment from Sec.~\ref{sec:multi-step}. We also find that if we set $\epsilon=0.25$ for Simple Set, the planner is grossly over-conservative and requires a step-wise help rate of $87\%$. 


\paragraph{Bimanual manipulation.} We additionally present results for a bimanual object rearrangement setup where ambiguities arise from the choice of the arm due to the limited reachability of each arm (\cref{fig:multi-step-prompt} right); results are deferred to \cref{app:bimanual}.


\subsection{Hardware: Mobile Manipulator in a Kitchen}
\label{subsec:saycan}

\setlength{\tabcolsep}{1pt}
\begin{wraptable}[9]{R}{6.5cm}
\vspace{-18pt}
\centering
\scriptsize
\begin{tabular}{cccccccc}\\\toprule  
Method & Model & $1-\epsilon$ & Plan Succ & Task Succ & Set Size & Help \\
\midrule
\knowno & PaLM-2L &
    0.85 & 0.87 & 0.76 & 2.22 & \bftab 0.67 \\
Simple Set & PaLM-2L &
    0.76 & 0.87 & 0.75 & 2.38 & 0.81 \\
No Help & PaLM-2L &
    - & 0.62 & 0.51 & - & 0 \\
\midrule
\knowno & PaLM-2L-IF &
    0.85 &0.86 & - & \bftab 1.86 & \bftab 0.67 \\
\knowno & GPT-3.5  &
    0.85 & 0.87 & - & 2.50 & 0.86 \\
\bottomrule
\end{tabular}
\caption{Results for \textbf{Hardware Mobile Manipulation}. Plan success rate is fixed between \knowno and Simple Set to compare the other metrics.
}
\label{tab:saycan}
\end{wraptable}

In this example, each scenario involves a mobile manipulator in front of a countertop and next to a set of recycling/compost/landfill bins in an office kitchen (\cref{fig:teaser}). The tasks include picking up some object from the counter, and possibly putting it in the drawer, or disposing of it in one of the bins. For the distribution of possible scenarios, we introduce new types of ambiguities based on Winograd Schemas \citep{levesque2012winograd} (\eg ``There is an apple and bottled water on the counter...it is rotten. Can you dispose of it?''), and ones that potentially involve unsafe actions (\eg ``place the bowl in the microwave."; there is a plastic bowl and a metal bowl, but only the plastic one is safe for the microwave). This is a single-step setting with multiple acceptable options. In \cref{tab:saycan}, we compare \knowno to Simple Set again by first setting $\epsilon=0.15$ and also finding $\epsilon=0.24$ for Simple Set that achieves the same plan success rate in simulation. The hardware experiment results again show that \knowno reduces the human help rate by $14\%$ and also reduces the average prediction set size. 


\paragraph{Target success guarantee from KnowNo is robust to varying LLM choice.} We also run \knowno with two other LLMs (without hardware evaluation). First, we use an instruction-finetuned version of PaLM-2L (PaLM-2L-IF); there is no significant performance difference from PaLM-2L; however, it generates smaller prediction sets in general by reducing the number of larger prediction sets (size 3 and 4) significantly. Second, we run GPT-3.5 (\textit{text-davinci-003}) from OpenAI. However, we find that it exhibits significant MCQA bias towards options D and E and against A and B, affecting the overall performance. Nonetheless, KnowNo still achieves $1-\epsilon$ target success rate, as the coverage guarantee from CP makes no assumption about the LLM confidences (e.g., calibrated or accurate) --- KnowNo flexibly compensates for the degraded LLM performance by triggering more human intervention.

\section{Related Work}
\label{sec:related work}

\paragraph{LLMs for robot planning and interaction.} Large language models have shown a wide range of capabilities: reasoning \cite{wei2022chain,kojima2022large}, logic \cite{suzgun2022challenging,creswell2022selection}, math \cite{lewkowycz2022solving}, physics  \cite{liu2022mind, ren2023tool}, high-level planning \cite{ahn2022can,huang2022language,xie2023translating,ding2023task,liu2023llm+,wu2023tidybot} with language feedback \cite{zeng2022socratic,huang2022inner}, and writing robot code \cite{liang2022code,singh2022progprompt,zelikman2023parsel}. 
The generated outputs can be guided to a certain extent with sufficient prompting, but LLMs are still prone to confidently hallucinating outputs (\eg referring to objects not observed in the scene \cite{zeng2022socratic}, or calling motion primitive APIs that may not exist \cite{liang2022code}). We hypothesize that these challenges can be alleviated while obtaining statistical guarantees by modeling the uncertainty of LLMs \citep{park2023clara} and generating prediction sets via CP. 

\paragraph{Uncertainty quantification for LLMs.} Motivated by LLMs' overconfidence and hallucinations, there has been a growing body of work in quantifying and better calibrating uncertainty \citep{zhou2023navigating, xiao2019quantifying, xiao2022uncertainty, kuhn2023semantic, kadavath2022language, lin2022teaching, mielke2022reducing}. 
In contrast to typical calibration methods that associate uncertainty with point-valued outputs, CP-based methods for language modeling provide coverage guarantees for set-valued predictors \citep{schuster2021consistent, schuster2022confident, deyInfilling2022, giovannotti2021transformer}. However, there have been few applications of CP in quantifying uncertainty of LLMs with \emph{free-form} outputs \citep{quach2023conformal}: \citet{kumar2023conformal} apply CP to next-token prediction in MCQA tasks exclusively, while \knowno builds on MCQA but is applicable to general natural language generation tasks.

\paragraph{Conformal prediction in robotics.}
To the best of our knowledge, this work is the first to employ CP for language-based planning. Prior work has utilized CP for fault detection, trajectory prediction, and planning in dynamic environments~\cite{luo2023sampleefficient, lindemann2022safe, lindemann2023conformal, dixit2022adaptive}. At each point in the planning horizon, the probabilistic safety guarantee either holds on average~\cite{dixit2022adaptive, luo2023sampleefficient}, or is too conservative due to union bounding~\cite{lindemann2022safe}, or requires additional calibration data to reduce conservatism~\cite{lindemann2023conformal}. In contrast, we provide a novel multi-step extension to CP to guarantee correctness for the entire planning horizon by performing sequence-level calibration in settings where the robot's actions influence the distribution of future inputs. 

\paragraph{Human-robot dialogue and interaction.} \knowno builds off previous work in robotics that addresses effective human-robot interaction through dialogue in both simulation and on hardware \cite{thomason2020cvdn, padmakumar2022teach, gao2022dialfred, banerjee2021robotslang}. \knowno uses a relatively simple setup between robot and human such that the human specifies potentially ambiguous instruction and clarifies it when the robot deems necessary; such setup does not consider human providing information related to robot observation or possible human error.
\section{Discussion}

\textbf{Summary:} We propose \knowno, a framework that applies conformal prediction (CP) to address the problem of uncertainty alignment for language-instructed robots, which we formalize as providing statistical guarantees of task completion while minimizing human help. Experiments across a variety of simulated and hardware setups demonstrate that \knowno achieves user-specified task completion levels consistently while reducing human help by $10-24\%$ 
compared to baseline approaches that lack formal assurances.


\textbf{Limitations and future work:} The primary limitation of our work is that the task completion guarantee assumes environments (objects) are fully grounded in the text input to the LLM, and the actions proposed by the LLM planner can be executed successfully. In the future, we are looking to incorporate uncertainty of the perception module (e.g., vision-language model) and the low-level action policy (e.g., language-conditioned affordance prediction) into the CP calibration. Another limitation is that, for the task guarantee to hold, the human needs to faithfully provide help when the robot needs it. Future work could also incorporate human modeling/error in the conformal prediction framework. Another exciting direction is to combine our methods with active preference learning \citep{sadigh2017active,biyik2022learning,brown2020better,myers2023active} to generate open-ended queries that maximally reduce uncertainty about human preferences. On the theoretical front, modifying CP to optimize different metrics for human help (e.g., minimizing human intervention rate by maximizing number of singleton sets) would be of practical interest. Overall, we hope that the work presented here spurs further efforts towards uncertainty alignment for safe and reliable language-instructed robots.



\acknowledgments{This work was partially supported by the NSF CAREER Award [\#2044149] and the Office of Naval Research [N00014-23-1-2148]. We thank Chad Boodoo for helping set up the UR5 hardware experiments, and Jensen Gao, Nathaniel Simon, and David Snyder for their helpful feedback on the paper.}
\clearpage



\small
\bibliography{bibliography}  
\normalsize

\renewcommand{\thetable}{A\arabic{table}}
\renewcommand{\theequation}{A\arabic{equation}}
\renewcommand\thefigure{A\arabic{figure}}
\renewcommand{\thesubsection}{A\arabic{subsection}}
\setcounter{figure}{0}
\setcounter{table}{0}
\setcounter{equation}{0}

\clearpage
\subsection{Evaluating Semantic Uncertainty of the LLM with MCQA}
\label{app:mcqa} 
Here we provide additional rationale of using the MCQA setup for evaluating LLM uncertainty. The uncertainty of the language model can be thought of as the predictive entropy of the output distribution. Consider the input tokens $x = (\sigma_i,\dots \sigma_k)$ and the output distribution $Y$ where $y = (\sigma_i,\dots \sigma_k) \in Y$:
\begin{equation}
    U(x) \ := H(Y | x) \ = \ - \int p(y|x) \ln p(y|x) dy.
\end{equation}
Evaluating this is very challenging for LLMs: the output distribution over $Y$ lies on the space of dimension $\mathcal{O}(|\mathcal{T}|^{k-i+1})$ where $\mathcal{T}$ is the set of possible tokens, and it has to be evaluated with a large number of samples and Monte-Carlo integration. Among the samples, there is also the bias against longer sequences \citep{murray2018correcting}.

We are partially inspired by \citet{kuhn2023semantic} that instead consider the \emph{semantic uncertainty} of the model: among samples in $Y$, there are groups of samples that have the same semantic meanings, such as ``put the sponge in the top drawer by first opening it'' and ``open the top drawer and put the sponge in it''. They may differ in $p(y|x)$ but we are not concerned with such uncertainty since it does not reflect the uncertainty of the LLM about the scenario. \citet{kuhn2023semantic} addresses this by first sampling a large number of samples from $Y$, and then grouping them based on some semantics classifier before evaluating the semantic uncertainty, which is the predictive entropy over the groups instead of over $Y$.

Could we improve the efficiency of finding semantically distinct groups in $Y$? The MCQA setup that we propose addresses this by prompting the LLM to generate likely, and also semantically different, options given the task using few-shot exemplars. 
We can think of this as splitting the output space $Y$into multiple spaces representing semantically different outputs. MCQA first samples the representative outputs from the spaces with higher weights in $Y$ (top four), and then includes the additional option ``an option not listed here'' to cover rest of the spaces. Unlike \citet{kuhn2023semantic} who calculate the entropy among the groups to decide whether to trust the answer to a question, \knowno instead combines the normalized probabilities with conformal prediction to provide set-based predictions with coverage guarantees.

\subsection{Proofs for CP in Multi-Step Setting}
\label{app:multi-step}

\begin{algorithm}[H]
\caption{Multi-step LLM planning with human help.}
\small
\begin{algorithmic}[1]
    \vspace{1mm}
    \For{time $t \leftarrow 0$ to $T-1$}
    \State Observe input $x_\text{test}^t$ and predict the set $C(x_\text{test}^t)$
    \If{$C(x_\text{test}^t)$ is a singleton}
        \State Execute the action in $C(x_\text{test}^t)$
    \Else
        \State Ask for human help
    \EndIf
\EndFor
\end{algorithmic}
\normalsize
\label{algo:adaptation}
\end{algorithm}


\noindent \textbf{Proof of Claim~\ref{claim:multi_CP}:} Suppose $\olsi{y} \in \olsi{C}(\olsi{x}_\text{test})$. We have, 
\begin{align}
    \olsi{y} \in \olsi{C}(\olsi{x}_\text{test})
     \iff &\min_t \ \hat{f}(x_\text{test}^t)_{y^t} \geq 1-\hat{q} &\\
    \iff &\hat{f}(x_\text{test}^t)_{y^t} \geq 1-\hat{q}, & \quad \forall t \in [T] \\
    \iff &y^t \in C^t(x_\text{test}^t), & \quad \forall t \in [T] \\
  \iff &\olsi{y}\in C(\olsi{x}_\text{test}).&
\end{align}

\noindent \textbf{Proof of Proposition~\ref{proposition: multi_CP}: }
Since we can bound the probability that $\olsi{y}_\text{test} \notin \olsi{C}(\olsi{x}_\text{test})$, we can also bound the probability that $\olsi{y}_\text{test} \notin C(\olsi{x}_\text{test})$. 
From the conformalization procedure, we have the following \textit{dataset-conditional} guarantee: with probability $1-\delta$ over the sampling of the calibration set $\olsi{Z}$, we have 
\begin{align}
    &\mathbb{P}\big(\olsi{y}_{\text{test}} \in \olsi{C}(\olsi{x}_{\text{test}}) | \olsi{Z}\big) \geq \text{Beta}^{-1}_{N+1-v , v} (\delta), \, \quad v= \lfloor(N+1)\hat{\epsilon}\rfloor \\
    \xRightarrow[]{\text{Claim~\ref{claim:multi_CP}}} &\mathbb{P}\big(\olsi{y}_{\text{test}} \in {C}(\olsi{x}_{\text{test}}) | \olsi{Z}\big) \geq \text{Beta}^{-1}_{N+1-v , v} (\delta),
\end{align}
where $\hat{\epsilon}$ is chosen such that $\epsilon = 1- \text{Beta}^{-1}_{N+1-v , v} (\delta)$. Hence, the following \textit{marginal} guarantee also holds:
\begin{align*}
    & \mathbb{P}\big(\olsi{y}_{\text{test}} \in \olsi{C}(\olsi{x}_{\text{test}})\big) \geq 1-\hat{\epsilon} \\
    \xRightarrow[]{\text{Claim~\ref{claim:multi_CP}}} &\mathbb{P}\big(\olsi{y}_{\text{test}} \in {C}(\olsi{x}_{\text{test}})\big) \geq 1-\hat{\epsilon}.
\end{align*}
This result provides a bound on the task completion rate if $\olsi{x}_\text{test}$ is drawn using the distribution $\mathcal{D}$. However, recall that the sequence $\olsi{x}$ of augmented contexts as defined in \cref{sec:multi-step} arises from having performed the \emph{correct} actions in previous steps; incorrect actions may result in a distribution shift. In order to obtain a bound on the task completion rate, we consider three cases at any given timestep: (1) the prediction set is a singleton and contains the correct label, (2) the prediction set is not a singleton but does contain the correct label, and (3) the prediction set does not contain the true label. The robot performs the correct action in the first two cases (without help in (1) and with help in (2)), while CP bounds the probability of case (3). Thus, the CP bound translates to a bound on the task success rate.  

As seen in~\cref{eq:minimality}, we have from \citep[Thm.~1]{sadinle2019least}, that we achieve the smallest average set size among all possible sequence-level prediction schemes, $\olsi{\mathcal{C}}$, if $\hat{f}$ models the prediction uncertainty accurately,
\begin{equation}
    \min_{\olsi{C} \in \olsi{\mathcal{C}}} \ \underset{(\olsi{x},\cdot) \sim {\mathcal{D}}}{\mathbb{E}} \big[|\olsi{C}(\olsi{x})|\big], \ \text{subject to} \ \mathbb{P}\big(\olsi{y} \in \olsi{C}(\olsi{x})\big) \geq 1-\hat{\epsilon}.
\end{equation} 

\subsection{CP in Settings with Multiple Acceptable Options Per Step}
\label{app:multi-option}
\begin{proposition}[Multi-label uncertainty alignment]\label{proposition: multi_label_CP}
Consider a setting where we use CP with coverage level $1-\epsilon$ to construct the prediction set when there are multiple true labels and seek help whenever the set is not a singleton at each timestep. With probability $1-\delta$ over the sampling of the calibration set, the task completion rate over new test scenarios drawn from $\mathcal{D}$ is at least $1-\epsilon$. 
\end{proposition}
\textbf{Proof: }We have a dataset of $Z = \{(\tilde{x}_i, Y_i),...\}_{i=1}^N$ sampled i.i.d. from a data distribution $\mathcal{D}$ for calibration (we use the same notation $\mathcal{D}$ as in the single-label setting here), where $Y_i := \{y_{i,j}\}_{j=1}^{J_i}$ is the set of true labels for a single trial. For each label, we use the same heuristic notion of confidence, $\hat{f}(x)_y \in [0,1]$.

We define an operator $\beta: \mathcal{X} \times \mathcal{Y}^J \rightarrow \mathcal{Y}$ where $\mathcal{X}$ is the space of contexts and $\mathcal{Y}$ is the space of labels:
\begin{equation}
    \beta(x, Y) := \arg \max_{y \in Y} \hat{f}(x)_y,
\end{equation} 
which takes the true label with the highest confidence value from the true label set.

If we consider applying $\beta$ to every point in the support of $\mathcal{D}$, a new distribution $\mathcal{D}'$ is induced. We also consider the induced dataset of samples $S' = \{ (x_i, y^{\max}_i) \}_{i=1}^N$, where $ y^{\max}_i := \beta(x_i, Y_i)$. Then we can perform the usual conformalization and obtain the guarantee that with
\begin{equation}
    C(x_\text{test}) := \{y | \hat{f}(x_\text{test})_y \geq 1-\hat{q}\},
\end{equation}
the following \textit{marginal} guarantee holds,
\begin{align}
    \mathbb{P}(y^{\max}_\text{test} \notin C(x_\text{test})) &\leq \hat{\epsilon}, \\
   \Rightarrow \mathbb{P}(\argmax_{y \in Y_\text{test}} \hat{f}(x_\text{test})_y \notin C(x_\text{test})) &\leq \hat{\epsilon}, \\
    \Rightarrow \mathbb{P}(\beta(x_\text{test}, Y_\text{test}) \notin C(x_\text{test})) &\leq \hat{\epsilon},
\end{align}

and the following \textit{dataset-conditional} guarantee holds when we choose $\hat{\epsilon}$ such that $\epsilon = 1-\text{Beta}^{-1}_{N+1-v , v} (\delta)$ where $v= \lfloor(N+1)\hat{\epsilon}\rfloor$,
\begin{align}
    \mathbb{P}(\beta(x_\text{test}, Y_\text{test}) \in C(x_\text{test}) | Z) &\geq 1-\epsilon .
\end{align}
Hence, $C(x_{\text{test}})$ contains the true label with the highest confidence with probability at least $1-\epsilon$. 

At test time, we sample $(x_\text{test}, Y_\text{test})$ from $\mathcal{D}$ that is i.i.d. with samples in $S$ ---  for the guarantee to hold for $\beta(x_\text{test}, Y_\text{test})$, we need to show $\beta(x_\text{test}, Y_\text{test})$ is a sample from $\mathcal{D}'$ that is i.i.d. with samples in $S'$. This is true since functions of independent random variables are independent, and functions of identically distributed random variables are identically distributed if the functions are measurable.



\subsection{CP in Multi-Step Setting with Multiple Acceptable Options Per Step}
\label{app:multi-step-multi-option}
\begin{proposition}[Multi-step, multi-label uncertainty alignment]\label{proposition: multi_step_label_CP}
Consider a multi-step setting where we use CP with coverage level $1-\epsilon$ to causally construct the prediction set when there may be multiple true labels at any step and seek help whenever the set is not a singleton at each timestep. With probability $1-\delta$ over the sampling of the calibration set, the task completion rate over new test scenarios drawn from $\mathcal{D}$ is at least $1-\epsilon$. 
\end{proposition}
\textbf{Proof: }For the multi-step setting, each trial now involves a sequence of contexts $\olsi{x}$ and a set of sequences of true labels:
\begin{equation}
    \olsi{Y} = \{ \olsi{y}_1, \olsi{y}_2, ..., \olsi{y}_{M} \},
\end{equation}
where $\olsi{y}_m := (y^0_m, y^2_m, ..., y^{T-1}_m)$. For example, $\olsi{Y}$ can contain the sequence of ``blue block, yellow block, green block'', ``green block, blue block, yellow block'', ..., for the task of picking up three blocks. We collect a dataset of $\olsi{Z} = \{(\olsi{x}_i, \olsi{Y}_i)\}$ of i.i.d. samples from the data distribution $\olsi{\mathcal{D}}$.

Unlike the single-step setting, here we cannot apply $\beta$ to the set of true labels in each step since we are reasoning over a \emph{set of sequences}, and not a \emph{sequence of sets} of true labels. Notably, the true label set at time step $t$ depends upon the sequence of previously chosen true labels.

Let $Y^t[x^0, \bar{y}^{t-1}]$ denote the set of true labels at timestep $t$, \emph{conditioned} upon the initial context $x^0$ and a partial sequence of past true labels $\bar{y}^{t-1} := (y^0, \ldots, y^{t-1})$ extracted from $\olsi{Y}$. We then autoregressively define the following sequence:
\begin{align}
    \olsi{\beta}_0(\olsi{x}, \olsi{Y}) &:= \argmax_{y \in Y^0} \hat{f}(x^0)_y, \quad Y^0 := \{y_1^0, \ldots, y_M^0\} \\
    \olsi{\beta}_t(\olsi{x}, \olsi{Y}) &:= \olsi{\beta}_{t-1}(\olsi{x}, \olsi{Y}) \ \bigcup \argmax_{y \in Y^t[x^0, \olsi{\beta}_{t-1}(\olsi{x}, \olsi{Y})]} \hat{f}(x^t)_y, \quad t = 1, \ldots, T-1.
\end{align}


For convenience, we denote $\olsi{\beta}_t(\olsi{x}, \olsi{Y})[\tau]$ the $\tau$ element in $\olsi{\beta}_t(\olsi{x}, \olsi{Y}), \tau \leq t$. An intuitive interpretation is that, we can consider $\olsi{Y}$ forming a tree of valid executions (all possible actions that can be taken by choosing each of true labels). 
Hence, at each time step $t$, $\olsi{\beta}_t(\olsi{x}, \olsi{Y})$ prunes the tree to a single branch by taking the true label with the highest heuristic value $\hat{f}(x^t)$. This reduces the tree of all possible sequences of true labels to a single branch of true labels with highest confidence. Given this single branch of true labels, we can now perform CP as shown in the multi-step setting in~\cref{app:multi-step}.


We apply $\olsi{\beta}_{T-1}$ to every point in the support of $\olsi{\mathcal{D}}$, and a new distribution $\olsi{\mathcal{D}}'$ is induced. We consider $\olsi{S}' = \{(\olsi{x}_i, \olsi{y}^{\max}_i)\}$, where $ \olsi{y}^{\max}_i := \olsi{\beta}_{T-1}(\olsi{x}_i, \olsi{Y}_i)$. Let $\olsi{Y}_\text{test}$ be the set of sequences of true labels for $\olsi{x}_\text{test}$. Suppose we get the \textit{marginal} bound with $\olsi{\beta}_{T-1}$ as the labels:
\begin{equation}
    \mathbb{P}(\olsi{\beta}_{T-1}(\olsi{x}_\text{test}, \olsi{Y}_\text{test}) \notin C(\olsi{x}_\text{test})) \leq \hat{\epsilon},
\label{eq:bound-multi-label-multi-step}
\end{equation}
and \textit{dataset-conditional} bound  when we choose $\hat{\epsilon}$ such that $\epsilon = 1-\text{Beta}^{-1}_{N+1-v , v} (\delta)$ where $v= \lfloor(N+1)\hat{\epsilon}\rfloor$,
\begin{equation}
    \mathbb{P}(\olsi{\beta}_{T-1}(\olsi{x}_\text{test}, \olsi{Y}_\text{test}) \notin C(\olsi{x}_\text{test})|\olsi{Z}) \leq \epsilon,
\label{eq:bound-multi-label-multi-step-conditional}
\end{equation}
which states that at test time, given a context sequence $\olsi{x}_\text{test}$, we produce a prediction set of sequences; if we consider a sequence consisting of the true label with the highest score at each step, the probability of this sequence covered by $C(\olsi{x}_\text{test})$ is lower bounded by $1-\epsilon$. However, we need to be careful of following $\olsi{\beta}_t$ at each step at test time. Conside the three cases:
\begin{itemize}[leftmargin=*]
    \item (1) At a given time-step, the prediction set $C^t(x^t_\text{test})$ does not contain the true label, $\olsi{\beta}_t(\olsi{x}, \olsi{Y})[t]$.
    \item (2a) The prediction set is a singleton and does contain the true label.
    \item (2b) The prediction set is not a singleton (but does contain the correct label).
\end{itemize}

We already bound the probability of (1) happening with the CP bound; (2a) is fine since the LLM will take the correct action; (2b) is more challenging --- in this case the robot asks the human for help, and we need to make sure the human ``follows'' the true label, by choosing the true label in the prediction set with the highest confidence by $\hat{f}$. In practice, we present the labels ranked by $\hat{f}$ and ask the human to choose the true label with the highest rank.

Now let's derive the bound in~\cref{eq:bound-multi-label-multi-step} and~\cref{eq:bound-multi-label-multi-step-conditional}. Again we need to consider the causal construction issue. As seen in Section~\ref{sec:multi-step}, we construct the prediction set $\olsi{C}(\olsi{x}_\text{test})$ non-causally using the score function $s_i = 1-\hat{f}(\olsi{x}_i)_{\olsi{y}^{\max}_i}$ (taking minimum over steps). For a test sequence $\olsi{x}_\text{test}$, we apply $\olsi{\beta}_{T-1}$ to the true label set of sequences $\olsi{Y}_\text{test}$ to get $ \olsi{y}^\text{max}_\text{test} = \olsi{\beta}_{T-1}(\olsi{x}_\text{test}, \olsi{Y}_\text{test})$. Now suppose $\olsi{y}^\text{max}_\text{test} \in \olsi{C}(\olsi{x}_\text{test})$, then we can show $\olsi{y}^\text{max}_\text{test} \in C(\olsi{x}_\text{test})$ with the same proof as the single-label setting, which gives us the bound.

Lastly we need to show the sampled test sequence from $\olsi{D}$ leads to a sample from $\olsi{\mathcal{D}}'$ i.i.d. with $\olsi{S}'$. This is true with the same argument that functions of independent random variables are independent.

\subsection{Additional Experiment Setting: Hardware Bimanual Setup}
\label{app:bimanual}

In this example, a bimanual setup with two Kuka IIWA 7 arms move objects on the table, with one bin at each side (\cref{fig:multi-step-prompt} right). The reachable workspace of each arm is limited so that one arm cannot reach the other end of the table or the other bin. Thus, there can be ambiguities in the choice of the arm depending on the task; \eg \cref{fig:multi-step-prompt} shows the human asking the robot to pass over the mango, but not specifying which side the human is standing at. \knowno is able to capture such ambiguities and triggers clarification. We design a scenario distribution (single-step with single acceptable options) with all instructions being ambiguous (thus requiring high human intervention rate): with $\epsilon=0.15$, the robot achieves 84$\%$ plan success with $92\%$ help. With 10 trials, the robot succeeds 9 times while triggering help for 9 times. Details of the scenario distribution are shown in \cref{app:experiment}.
\subsection{LLM Prompt Setup}
\label{app:prompt}

Next we detail the LLM prompt setup for MCQA applied in \knowno. We will use Mobile Manipulation from \cref{subsec:saycan} as the example. 

\paragraph{Multiple choice generation.} Given a scenario, we first prompt the LLM to generate four semantically different options for possible next step. We apply few-shot prompting as shown in \cref{fig:app-mc-pre-prompt} below. In this scenario, there is a Coke, a bottled tea, and a Pepsi on the counter, and the task is to put the Coke in the top drawer but the choice of drawer is under-specified (``Put the Coke in the drawer please.'').

\begin{figure}[H]
\begin{center}
\includegraphics[width=0.9\textwidth]{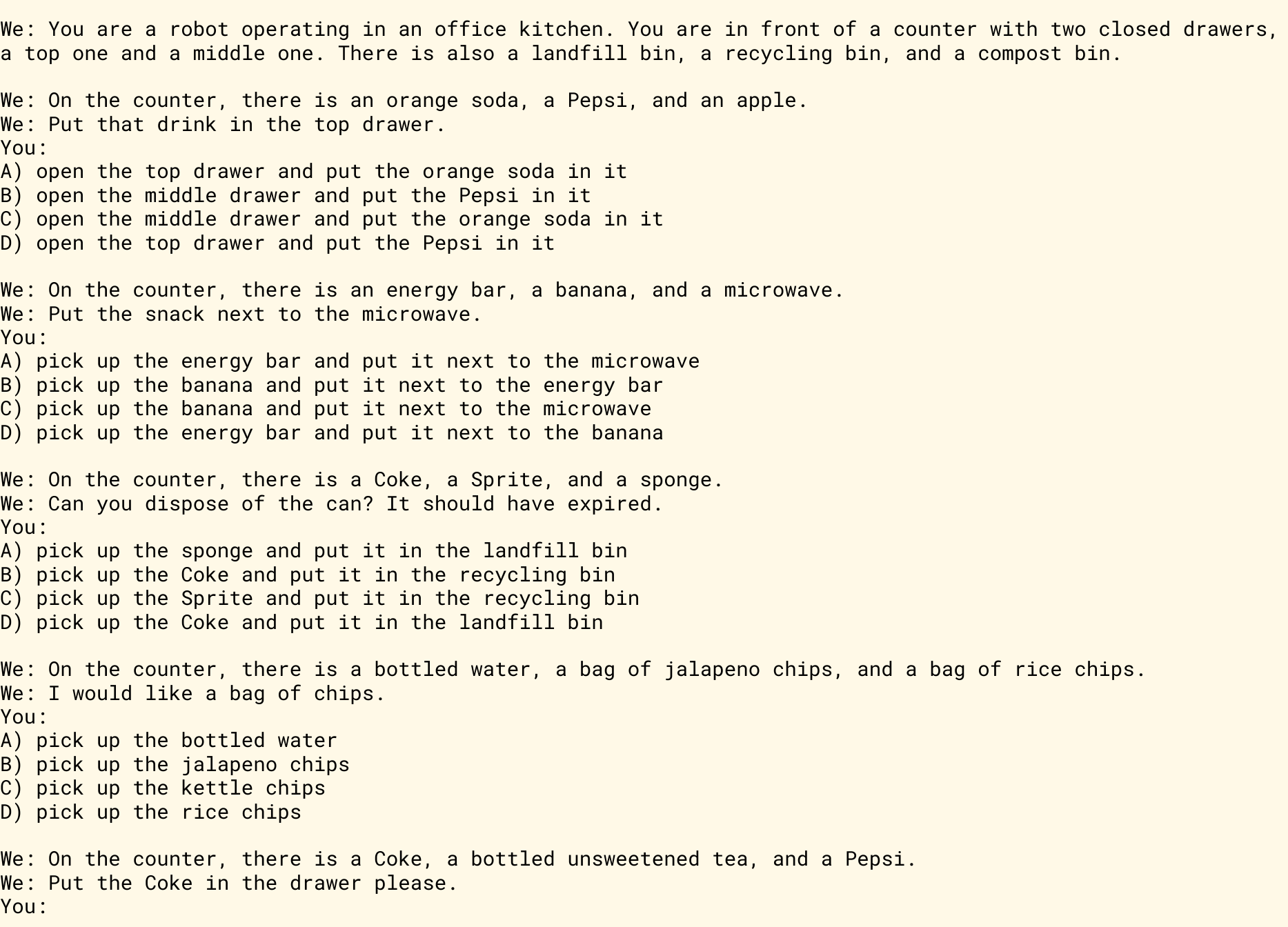}
\caption{Prompt used for multiple choice generation in Mobile Manipulation. With few-shot prompting show examples of possible next steps in different scenarios, the LLM can generate \emph{semantically different} plans that are more likely than others (see prompting result in \cref{fig:app-mc-post-prompt}).}
\vspace{-10pt}
\label{fig:app-mc-pre-prompt}
\end{center}
\end{figure}

\begin{figure}[H]
\begin{center}
\includegraphics[width=0.9\textwidth]{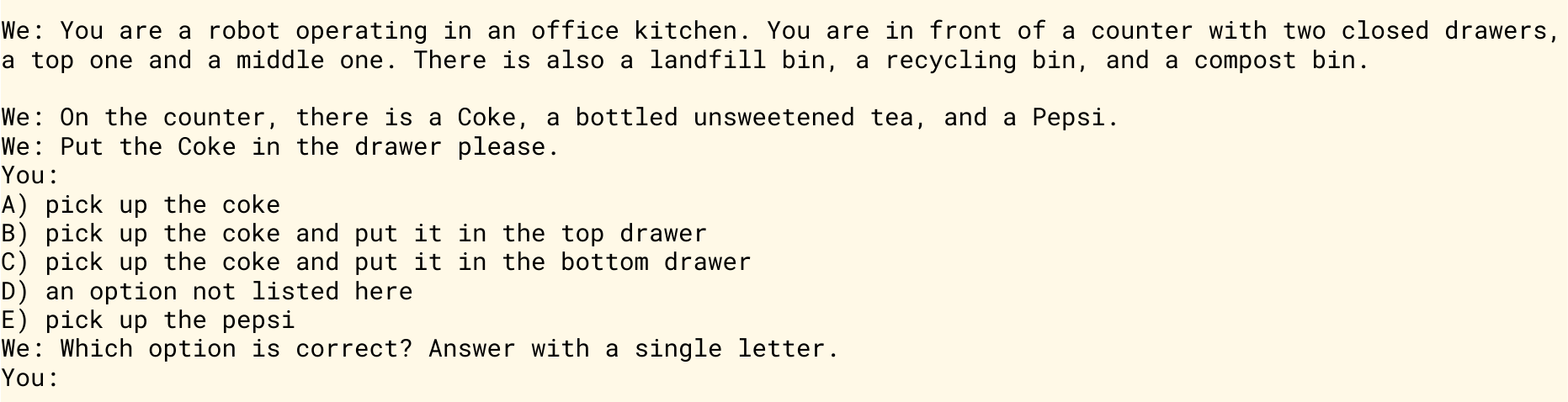}
\caption{Prompt used for next-token prediction with generated multiple choices in Mobile Manipulation.}
\vspace{-10pt}
\label{fig:app-mc-post-prompt}
\end{center}
\end{figure}

After the LLM generates four options, we append an additional option `an option not listed here' to the four generated ones and then randomize the order to further prevent bias. We then use a zero-shot prompt in \cref{fig:app-mc-post-prompt} for querying next-token probabilities (`A', `B', `C', D', `E').

\subsection{Additional Experiment Details}
\label{app:experiment}

\begin{figure}[h]
\begin{center}
\includegraphics[width=0.8\textwidth]{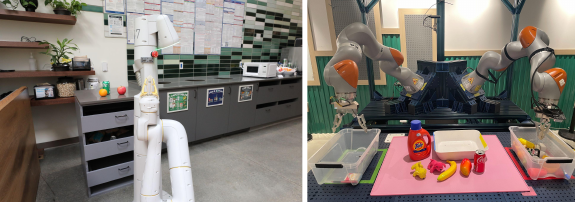}
\caption{(Left) Office kitchen environment with drawers and bins for Mobile Manipulation. (Right) Bimanual setup with the set of objects used in the experiments.}
\vspace{-10pt}
\label{fig:app-envs}
\end{center}
\end{figure}

\paragraph{Environments.} In addition to \cref{fig:teaser} and \cref{fig:multi-step-prompt}, here \cref{fig:app-envs} shows the office kitchen environment with the set of drawers and bins used in Mobile Manipulation (left), and the bimanual setup with the set of objects used on the mat used in Bimanual (right). There is another set of drawers used in the mobile manipulation experiments underneath a much bigger countertop not shown here.

\textbf{Scenario Distribution and Calibration Dataset} Next, we provide more details on the parameterization of the scenario distribution in each experiment setting, in particular, the possible ambiguities with the instruction and goal. With the distributions set up, the calibration dataset are then generated by randomly sampling 400 i.i.d. scenarios from them:
\begin{itemize}[leftmargin=10pt]
    \item Simulated setting: 
    \begin{itemize}
        \item Environment: there are always three blocks and bowls of color red, yellow, and green with random locations on the table.
        \item Goal: we use the following template: \ \{put,\ place,\ move\} \ \{a,\ one,\ a single of,\ two,\ a pair of,\ three,\ all,\ red,\ yellow,\ green\} \ \{block(s), \ bowl(s)\} \ \{on, \ to the left of, \ to the right of, \ to the front of, \ at the back of\} the \ \{red, \ green, \ yellow\} \ \{block(s), \ bowl(s)\}. The scenario distribution is uniform over these possibilities.
        \item Instruction: for the language instructions, we modify the goal (from the template) according to the ambiguity type. The scenario distribution is uniform over the listed ambiguous cases in each ambiguity type.
        \begin{itemize}
            \item Attribute ambiguities: refer to the block as one of ``cube'', ``cuboid'', ``box'',  ``square object'', to the bowl as one of ``container'', ``round object'', ``receptacle'', or to either block or bowl as one of ``object'', ``item'', ``thing'' (``move the blue object in yellow bowl''); refer to ``blue'' as one of ``cyan'', ``navy'', to ``green'' as one of ``greenish'', ``grass-colored'', and to ``yellow'' as ``orange'' or ``gold''. This setting is the least ambiguous one among the three ambiguity types.
            \item Numeric ambiguities: refer to either two or three numerically with one of ``a few'', ``a couple of'', ``some'', ``a handful of'' (``put some blocks in the green bowl'').
            \item Spatial ambiguities: refer to any of the four possible directions with ``near'', ``close to'', ``beside'', ``next to'', refer to either left to right with ``lateral to'', and refer to either front or behind with ``along the line of sight''. This setting is the most ambiguous one among the three ambiguity types.
        \end{itemize}
    \end{itemize}
    \item Hardware Tabletop Rearrangement setting: 
    \begin{itemize}
        \item Environment: there are three items to be sorted placed randomly on the table, and there is a blue plate and a green plate. 28 toy items are split (\cref{fig:app-detection}) into two categories of human liking them or disliking them: the things the human likes include corn, avocado, celery, carrot, tomato, lettuce, apple, orange, pear, lemon, peanut butter, sunny-side-up egg, egg, and pea; the human dislikes pretzel, cracker, waffle, mustard, ketchup, pizza, meat patty, cheese, chicken drumstick, peach, mango M\&M, Skittles, and donut. The three items on the table are uniformly sampled ones from the two categories (1 from likes and 2 from dislikes, or 2 from likes and 1 from dislikes).
        \item Goal: place the items that human likes in the blue plate, and ones that human dislikes in the green plate.
        \item Instruction: to make the goal ambiguous, only a subset of human preferences are revealed in the prompt:``I like: corn, celery, pear, peanut butter, sunny-side-up egg. I don't like: pretzel, cracker, mustard, peach, M\&M''. And then the instruction includes ``Can you sort things I like and dislike in blue and green plates?''
    \end{itemize}
    \item Hardware Mobile Manipulator setting: 
    \begin{itemize}
        \item Environment: the full list of possible objects include: bottled water, bottled tea, orange soda, RedBull, Coke, Pepsi, Sprite, rice chips, jalapeno chips, kettle chips, multigrain chips, apple, orange, energy bar, clean sponge, dirty sponge, metal bowl, plastic bowl. Depending on the ambiguity listed below, there is three objects placed on the top of the counter (including randomly sampled distractors from the list). There is also a set of landfill, compost, and recycling bins, a microwave, and a portable stove.
        \item Instruction: for convenience, we introduce the possible instructions first in different ambiguous scenarios; they each correspond to possible goals. Please refer to \url{https://robot-help.github.io/prompts/mobile_tasks.txt} for the full list. The possible instructions are a uniform distribution over different types: (1) single-label, \eg `Bring me a Coke' (unambiguous); (2) creative-single-label, \eg `I want a healthy fruit to munch on.' which means the apple (unambiguous); (3) multi-label, \eg `Bring me a cola.' and either Coke or Pepsi is acceptable; (4) creative-multi-label, \eg `Bring me something with a kick.' and either RedBull or jalapeno chips are acceptable; (5) spatially-ambiguous, \eg `Put the Coke in the drawer' or `Put the Coke near the fruit' which under-specifies the drawer or fruit; (6) unsafe, \eg `Can you dispose of the bottle drink? It should have expired.' or `Place the bowl on the stove, please.'; (7) Winograd, \eg 'There is a sponge and a bag of rice chips...I don't want to use it for cleaning any more. Can you please dispose of it?" We use the GPT-4 model for generating the creative tasks.
        \item Goal: the corresponding goal for the ambiguous instructions above. For example, the instruction is ``Put the Coke in the drawer'', and the goal is uniform over the two possibilities: put the Coke in the top drawer, and put the Coke in the bottom drawer.
    \end{itemize}
    \item Bimanual setting: 
    \begin{itemize}
        \item Environment: the full list of possible objects include: Coke can, Sprite can, green apple, banana, mango, red block, green block, yellow block, pink plushie, yellow plushie, purple plushie. Depending on the ambiguity listed below, there is three objects placed on the table (including randomly sampled distractors from the list). There is possibly a human standing at the left or right side of the table.
        \item Instruction: Please refer to \url{https://robot-help.github.io/prompts/bimanual_tasks.txt} for the full list of scenarios sampled from the distribution. The possible instructions are a uniform distribution over different types: (1) `Pick up the \ \{object\} and pass it to me. I am next to the bin.' (2) `Pick up the \ \{object\} with the left arm.' (3) `Put the \ \{object\} in the bin closer to it.' (4) `Pick up the \ \{object\} with the arm closer to it.' (5) `Pick up the \ \{object\}.' (6) `Pick up the \ \{object\} at the handle.' (7) `Move the \ \{object\} to the front of the table.' (8) `Move the \ \{object\} on the sticky rubber mat to the front of the table.'
        \item Goal: the corresponding goal for the ambiguous instructions above. For example, the instruction is ``Pick up the Coke can and pass it to me. I am next to the bin.'', and the goal is pick up the Coke can with the left arm and pass it to human, if human is at the left side, and with the right arm if human is at the right side.
    \end{itemize}
\end{itemize}

\begin{figure}[h]
\begin{center}
\includegraphics[width=0.9\textwidth]{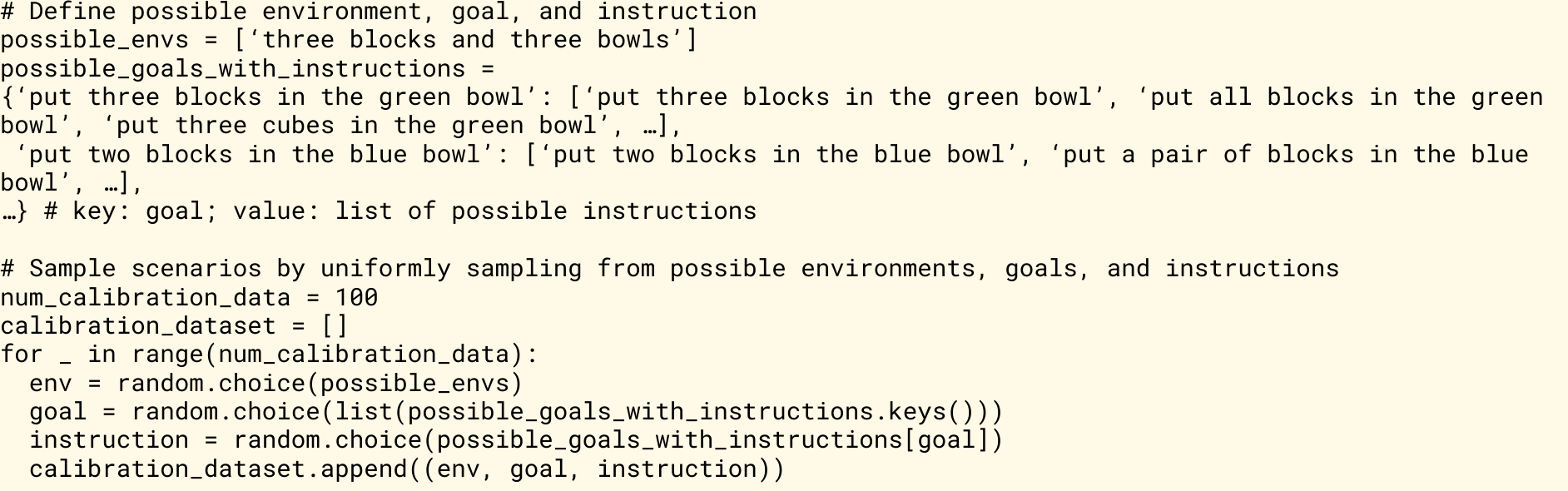}
\caption{Sample algorithm for defining the scenario distribution and generating the calibration dataset in the Simulation setting.}
\vspace{-10pt}
\label{fig:pseudocode}
\end{center}
\end{figure}

Next we provide more details on the baselines that require additional prompting strategies.

\textbf{Baselines - Ensemble Set.} Our ensemble-based method is a weaker method than the traditional model-based ensemble where multiple copies of neural network are trained and inferred with; however, this is infeasible with the LLM we use. In our work, we randomize over the few-shot examples in the prompt as the ensemble. We select a pool of 20 possible MCQA examples (see examples in Fig.~\ref{fig:app-mc-pre-prompt}), and then randomly sample a certain amount from it for each inference. Note that in this case, Ensemble Set actually has advantage over \knowno and Simple Set that, for the same data, it has seen many more examples than the fixed ones in the prompt used in \knowno and Simple Set. We only apply ensemble for next-token prediction; the same set of multiple choices generated is used.

\begin{figure}[h]
\begin{center}
\includegraphics[width=0.9\textwidth]{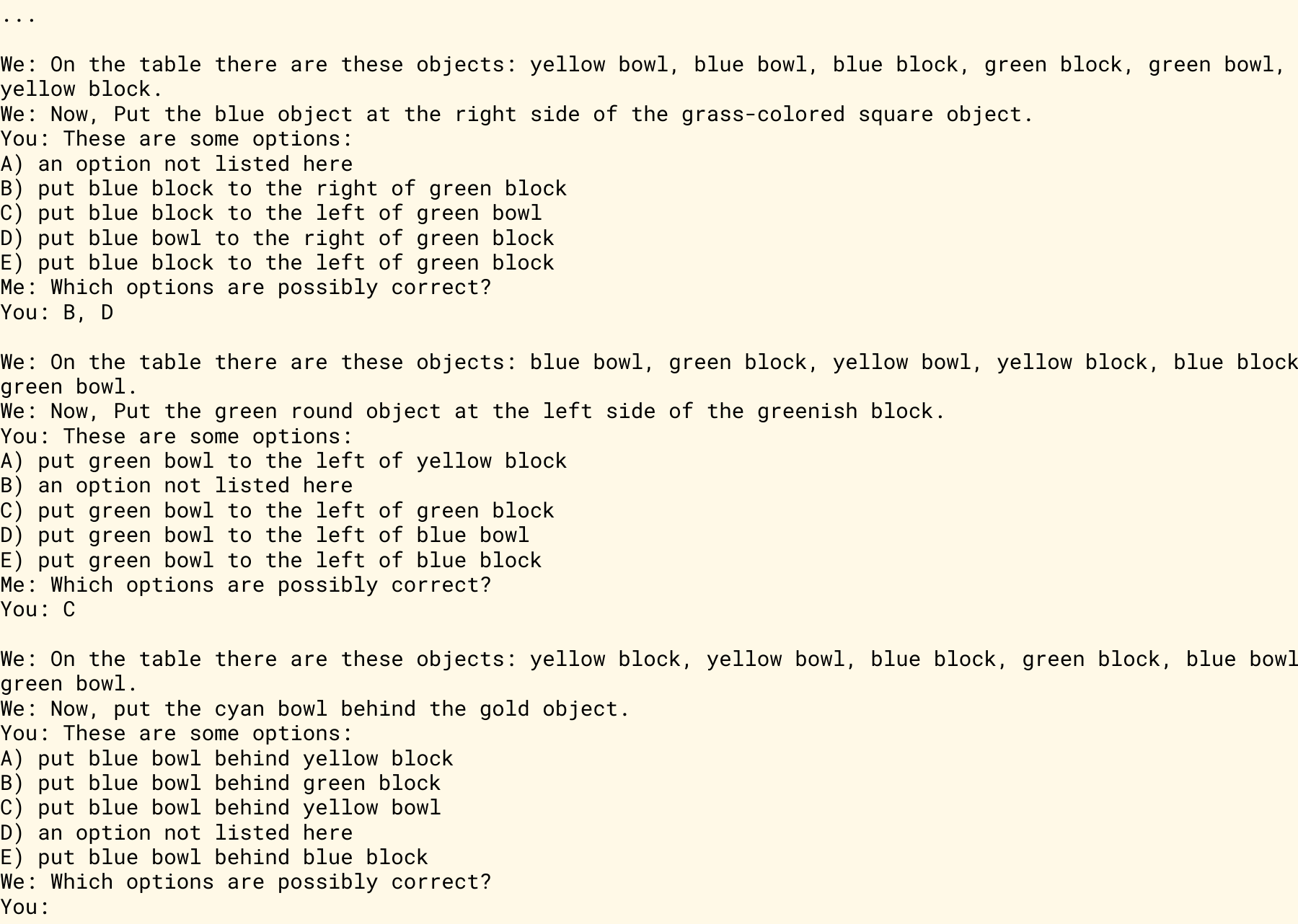}
\caption{Prompt with few-shot examples (two shown) for LLM generating the prediction set in Prompt Set baseline.}
\vspace{-10pt}
\label{fig:app-prompt-set}
\end{center}
\end{figure}

\textbf{Baselines - Prompt Set.} First, multiple choices are generated in the same way as \knowno. Then LLM is prompted to generate the prediction set, with few-shot examples in the prompt showing the possible labels (\cref{fig:app-prompt-set}). For example, ``We: Which options are possibly correct? You: A, C, D.''.

\begin{figure}[h]
\begin{center}
\includegraphics[width=0.9\textwidth]{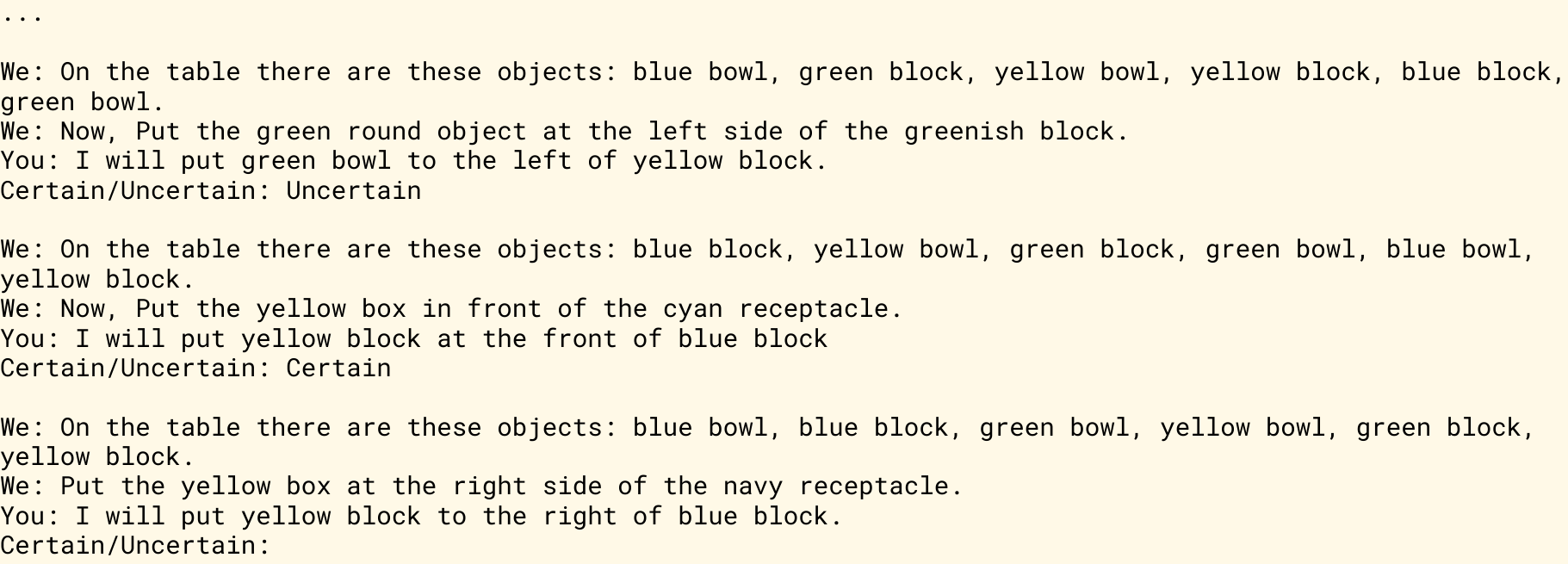}
\caption{Prompt with few-shot examples (two shown) for LLM expressing binary uncertainty in Binary baseline.}
\vspace{-10pt}
\label{fig:app-binary}
\end{center}
\end{figure}

\textbf{Baselines - Binary.} Instead of generating multiple choices, the LLM is first prompted to give the most likely action (\eg ``We: Put the Coke can in the drawer. You: I will'' shown in \cref{fig:app-perplexity-prompt}). Then we attach the generated response to the same prompt, and ask LLM to label ``Certain/Uncertain:'' given few-shot examples (\cref{fig:app-binary}).

\subsection{Additional Implementation Details}
\label{app:implementation}


While the focus of \knowno is mainly on providing uncertainty alignment for the LLM-based planner, below we provide details of the perception and action modules applied in all examples.

\begin{figure}[h]
\begin{center}
\includegraphics[width=0.8\textwidth]{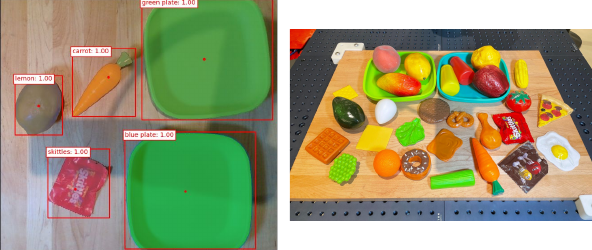}
\caption{(Left) MDTER \citep{kamath2021mdetr} object detection with Segment Anything \citep{kirillov2023segany} and most distant internal point (red dots) for Hardware Tabletop Rearrangement. (Right) The total 28 toy items used for the experiments.}
\vspace{-10pt}
\label{fig:app-detection}
\end{center}
\end{figure}

\paragraph{Perception.} For all examples except for the Mobile Manipulation, we use either MDETR \citep{kamath2021mdetr} (Hardware Tabletop Rearrangement) or Owl-ViT \citep{minderer2022simple} (Simulation and Bimanual) open-vocabulary object detector for recognizing the objects in the environment and obtaining the object locations for low-level action. In Simulation and Bimanual, the variations of the object types are limited, and with general prompting, the objects are detected without issue. In Hardware Tabletop Rearrangement, since we are use a wide variety of toy items (\cref{fig:app-detection} right), the detector has issues often differentiating objects like peanut butter and meat patty that are both darker colors. We modify the scenario distributions to avoid using such items together in one scenario. In addition, we apply the Segment Anything model \citep{kirillov2023segany} to extract the object segmentation masks (shown overlaid in \cref{fig:app-detection} left), and then use the \texttt{polylabel} algorithm \citep{Agafonkin_Polylabel_a_fast_2016} to find the most distant internal point of the mask as the suction point (shown as red dots).

\paragraph{Low-level action.} In Simulation and Hardware Tabletop Rearrangement, simple pick-and-place actions are executed based on object locations and solving the inverse kinematics. In Bimanual, the reachability of the Kuka arm is limited, and the pick-and-place action trajectories are solved using Sequential Quadtratic Programming (SQP) instead \citep{singh2022optimizing}. In Mobile Manipulation, for most of the tasks that involve simple pick-and-place and opening the drawers, the action is from an end-to-end policy from the RT-1 policy (please refer to \citep{brohan2022rt} for details), which takes in the raw observation. For some of the hard tasks such as putting the plastic bowl in the microwave and putting the metal bowl on the bowl, object locations are assumed known and we use scripted action policies.

\paragraph{Human feedback.} In \knowno, once human help is triggered, human is presented with the prediction set to choose the correct action from (if there is one). For example, the prediction set could include `A) put peanut butter in blue plate' and `C) put peanut butter in green plate' in Hardware Tabletop Rearrangement. In practice, we can convert the prediction set to a question in more natural language, \eg ``Do you like peanut butter or not?'' using simple heuristics. In Mobile Manipulation and Bimanual, we prompt the LLM to generate the question based on the prediction set.

\subsection{Additional Discussions}
\label{app:discussions}

\begin{figure}[h]
\begin{center}
\includegraphics[width=0.9\textwidth]{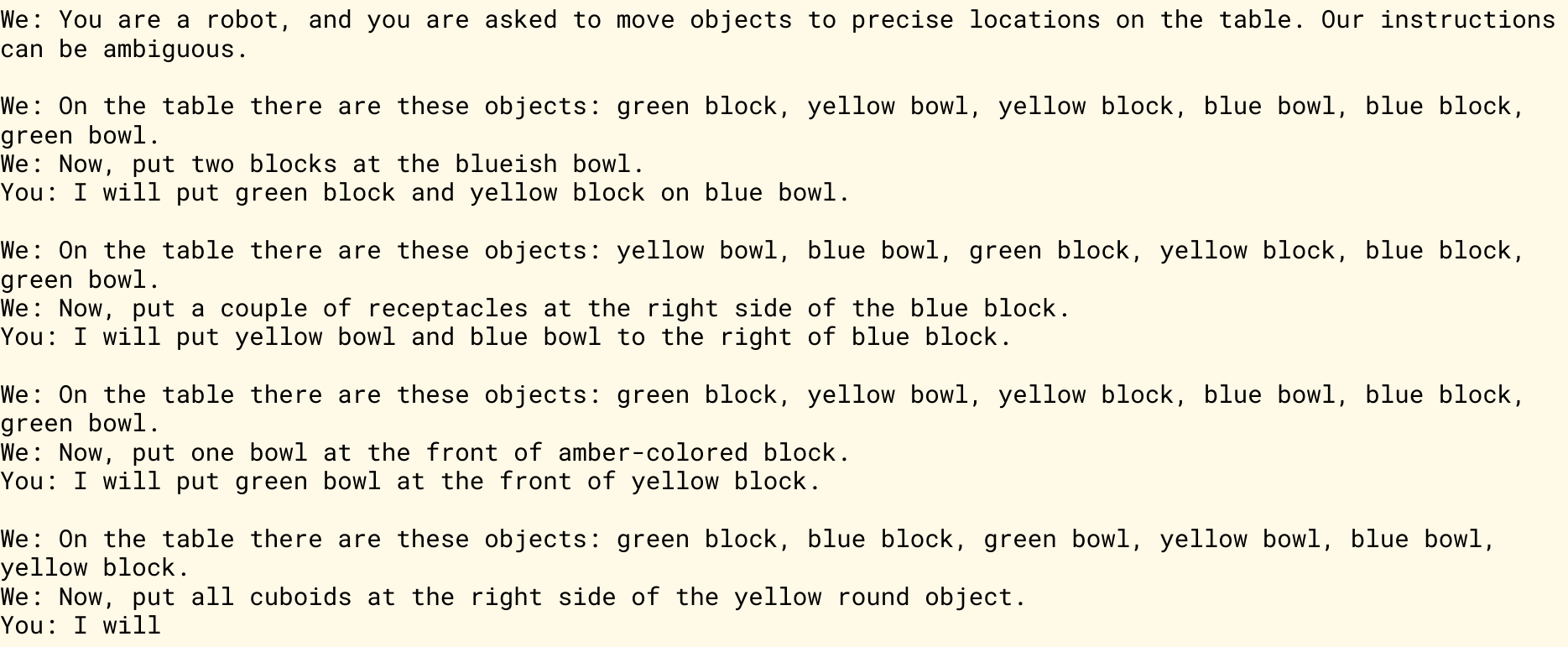}
\caption{Prompt used when evaluating the sentence-level scores in Simulation. We first generate the multiple choices in the same way as \knowno, and then evaluate the score of each multiple choice with this prompt. This prompt is also used to generate the option in the Binary baseline.}
\vspace{-10pt}
\label{fig:app-perplexity-prompt}
\end{center}
\end{figure}

\paragraph{Sentence-level score leads to worse performance.} In \cref{sec:overview} we hypothesize that the distribution of probabilities of LLM outputs $p(y)$ is highly sensitive to the output length. Here we explore the effect of using sentence output and the perplexity score for CP in Simulation. We still apply multiple choice generation first to obtain the possible options from LLM, and then query LLM scoring, for example, the probability of ``put the blue block in the green bowl'' with the prompt ending with ``I will'' (\cref{fig:app-perplexity-prompt}). We test two different method: CP with RawProb, which uses the raw product of conditional probabilities $p(y)=\prod_{i=1}^{k} p(\sigma_i \mid \sigma_1, \ldots, \sigma_{i-1})$, and CP with Perplexity, which applies length normalization and commonly used in language modeling, $\textit{perplexity}(y)=\exp\{-\frac{1}{k}\sum_{i=1}^{k} \log \ p(\sigma_i \mid \sigma_1, \ldots, \sigma_{i-1})\}$. \cref{tab:perplexity-comparison} shows that for all three settings, using either sentence-level score leads to worse performance, and performance degradation correlates with variance of the multiple choice lengths. We also notice that there is no significant difference in performance between CP with RawProb and CP with Perplexity; this could indicate that beyond eliminating length bias, applying MCQA setup in \knowno affords additional benefit of soliciting more accurate confidence from the LLM.

\begin{table}[H]
\vspace{-10pt}
\centering
\footnotesize
\begin{tabular}{ccccc}\\\toprule
Setting & Multiple Choice Length Variance & Method & Set Size & Help \\
\midrule
\multirow{3}{*}{Attribute} & \multirow{3}{*}{1.52}
    & \knowno & 1.18 & 0.18 \\
    &
    & CP w/ RawProb & 1.33 & 0.32 \\
    &
    & CP w/ Perplexity & 1.35 & 0.36 \\
    \midrule
\multirow{3}{*}{Spatial} & \multirow{3}{*}{2.81}
    & \knowno & 2.23 & 0.69 \\
    &
    & CP w/ RawProb & 2.50 & 0.82 \\ 
    &
    & CP w/ Perplexity & 2.67 & 0.88 \\ 
    \midrule
\multirow{3}{*}{Numeric} & \multirow{3}{*}{8.51}
    & \knowno & 2.17 & 0.79 \\
    &
    & CP w/ RawProb & 4.06 & 1.00 \\ 
    &
    & CP w/ Perplexity & 4.06 & 1.00 \\ 
\bottomrule
\end{tabular}
\caption{Comparison of \knowno with CP with sentence output and perplexity score in the three ambiguity settings in Simulation. $\epsilon=0.15$. See \cref{fig:app-sample-mc} for samples of multiple choices generated with varying lengths.
}
\label{tab:perplexity-comparison}
\end{table}
\vspace{-15pt}

\begin{figure}[h]
\begin{center}
\includegraphics[width=0.9\textwidth]{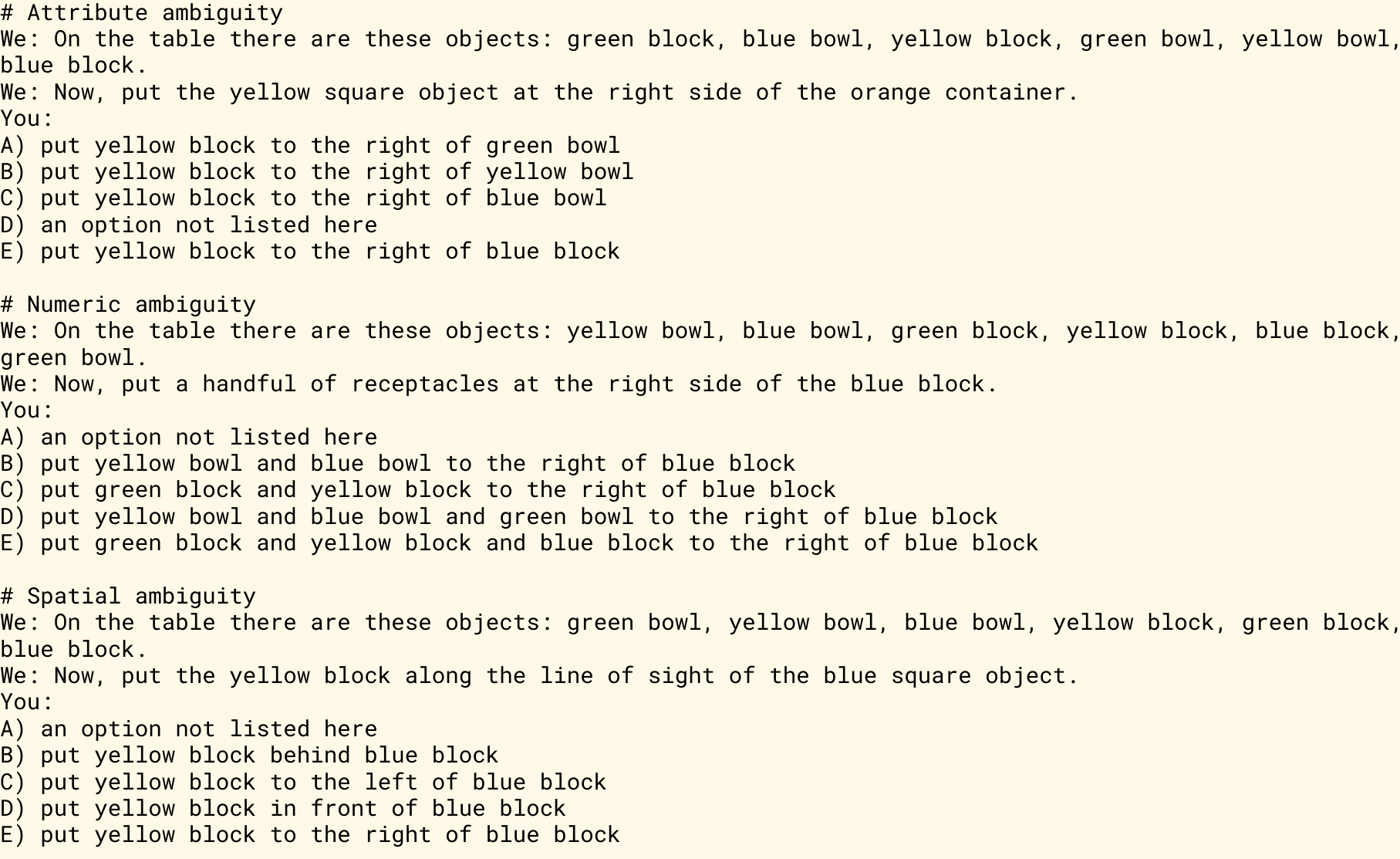}
\caption{Sample multiple choices generated with Attribute, Numeric, and Spatial ambiguities in Simulation.}
\vspace{-15pt}
\label{fig:app-sample-mc}
\end{center}
\end{figure}

\textbf{Potentially stronger baselines with model fine-tuning.} In \cref{sec:experiments} we introduce the two prompt-based baselines Prompt Set and Binary, and demonstrate them being (1) inflexible (not allowing controlling the target success rate) and (2) do not properly model the uncertainty. We note that these two baselines can be potentially strengthened by fine-tuning the LLM to better predict the binary uncertainty or the uncertainty set, if the true labels can be properly defined. In fact, some recent work \citep{lin2022teaching, kadavath2022language} have explored model fine-tuning and exhibiting the effectiveness of Binary for uncertainty calibration. We also explored fine-tuning the GPT3 model (\textit{davinci}) from OpenAI, which is the most powerful one from OpenAI available for fine-tuning. However, we find the model performing at very low accuracy with MCQA, and fine-tuning the model always results in overfitting to the dataset, even with thousands of data and varying hyperparameters (including ones from \citep{lin2022teaching} and default ones from the API). We suspect that our scenarios exhibit high complexity and variance, and it is non-trivial to fine-tune the model well with our dataset. Nonetheless, we hope to have future work looking into better training the model for proper uncertainty, and then applying CP on top of it to achieve set-based calibration.

\textbf{Low-level control success rate.} \knowno translates the coverage guarantee from CP to task completion guarantee leveraging human help. However, this relies on the low-level control working reliably. In Simulation Tabletop Rearrangement, we find the pick-and-place primitives always executed as the object diversity is limited to square blocks and normal bowls (only differing in color). In Hardware Tabletop Rearrangement, we find the pick-and-place primitive only failed once during the 50 trials of running \knowno and twice for Simple Set (\cref{tab:tabletop-multi-step}). The high success rate is largely thanks to the precise object masks from Segment Anything \citep{kirillov2023segany}. Also, to allow reliable suctioning, we apply clear scotch tape on some of the objects (\eg donut, waffle) to smoothen the surfaces. In Hardware Mobile Manipulation, we find the low-level action success rate to be around 86\%, which causes the non-trivial discrepancies between plan success and task success rates in \cref{tab:saycan}. One exciting future direction is to quantify and better calibrate the uncertainty of the low-level action module, and take such uncertainty into account of the end-to-end task completion guarantee.

\begin{figure}[h]
\begin{center}
\includegraphics[width=0.62\textwidth]{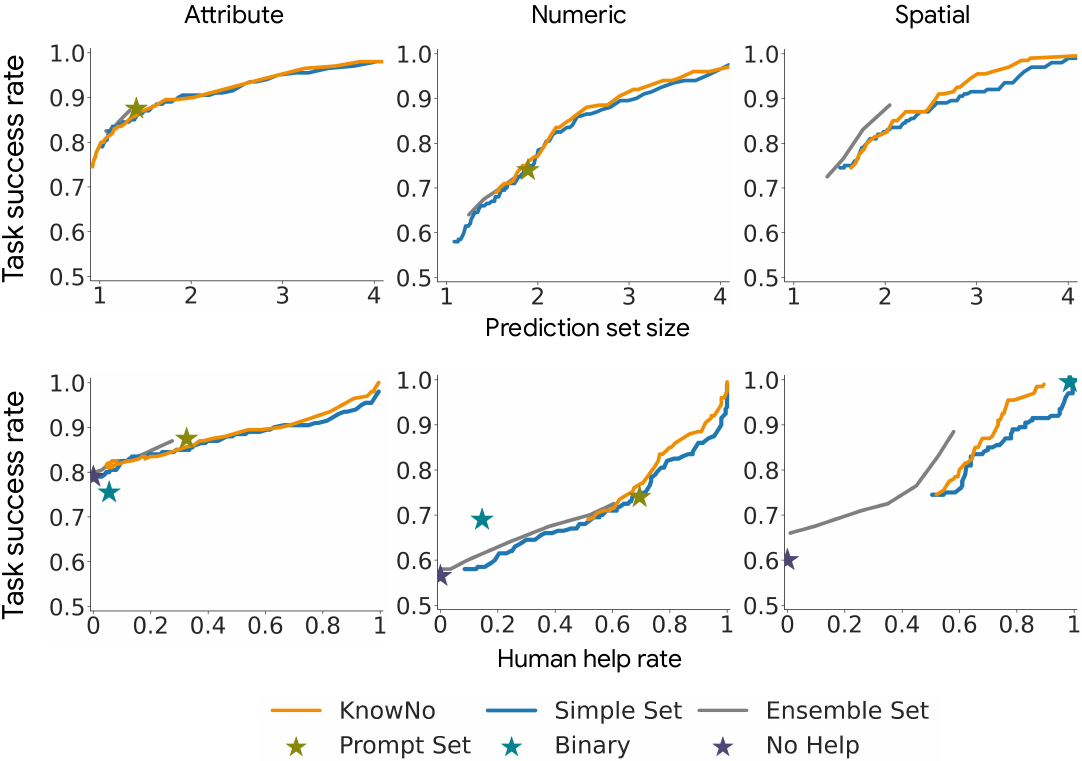}
\caption{Comparison of task success rate vs average prediction set size (Top row) and vs. human help rate (Bottom row) in Simulation for each of the three ambiguity settings: Attribute, Numeric, and Spatial (Columns), as $\epsilon$ varies from 0.25 to 0.01 for \knowno, and from 0.6 to 0.01 for Simple Set and Ensemble Set. Binary and No Help are not shown in the top row since prediction set is not involved. Prompt Set does not show up in the human help rate plot for the Spatial setting as it only achieves 35\% task success. As the setting gets more ambiguous (least in Attribute and most in Spatial), the performance difference between \knowno and Simple Set grows.}
\vspace{-10pt}
\label{fig:app-sim-comparisons-setting}
\end{center}
\end{figure}

\begin{figure}[h]
\begin{center}
\includegraphics[width=0.9\textwidth]{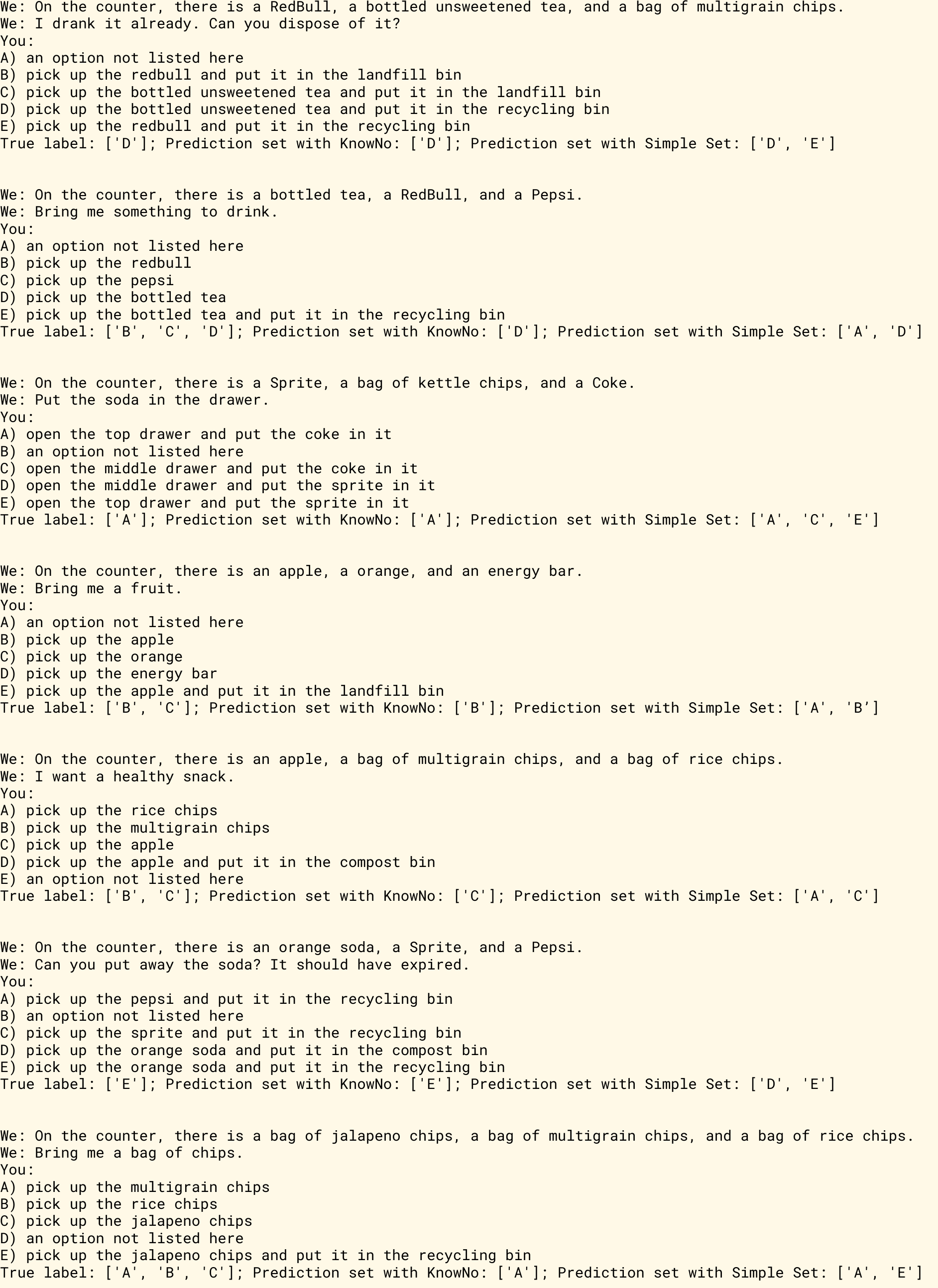}
\caption{Randomly sampled scenarios from Mobile Manipulation, where \knowno generates singleton prediction sets that contains one of the true labels and avoids human intervention, while Simple Set generates non-singleton prediction sets and asks for human help. In \cref{tab:saycan} we show that \knowno reduces the human intervention rate by $14\%$. We also point out the third example where the instruction is to `Put the soda in the drawer', which is ambiguous in the choice of top and middle drawer. In this scenario, it happens that the human means the top drawer, and \knowno generates the prediction set that only include the option `open the top drawer and put the coke in it'. This example exhibits the inherent bias in LLM (\eg bias towards top drawer over middle drawer).}
\vspace{-10pt}
\label{fig:app-sample-comparison}
\end{center}
\end{figure}



\end{document}